\pdfoutput=1
\documentclass{article}
\usepackage{iclr2027_conference}
\usepackage[T1]{fontenc}
\renewcommand{\sfdefault}{phv}

\usepackage{amsmath,amsfonts,bm}

\def\eqref#1{equation~\ref{#1}}

\def\1{\bm{1}}

\DeclareMathAlphabet{\mathsfit}{\encodingdefault}{\sfdefault}{m}{sl}
\SetMathAlphabet{\mathsfit}{bold}{\encodingdefault}{\sfdefault}{bx}{n}

\usepackage{hyperref}
\hypersetup{hidelinks}
\usepackage{url}
\usepackage{graphicx}
\usepackage{booktabs}
\usepackage{array}
\usepackage{float}
\usepackage{microtype}
\usepackage{placeins}
\usepackage{tabularx}
\usepackage{tikz}
\usetikzlibrary{arrows.meta,positioning,fit}
\usepackage{xspace}

\title{VStress: Correlation-Aware Auditing and\\Adaptive Budget Allocation\\for Repeated Verifiers}
\author{Miaobo Hu$^{1,2,*}$, Shuhao Hu$^{2}$, Xiaobo Guo$^{2}$, Xin Wang$^{2}$,\\
Bokun Wang$^{2}$, Peng Zhang$^{2}$, Daren Zha$^{2}$\\[4pt]
{\normalfont $^{1}$School of Artificial Intelligence, University of Chinese Academy of Sciences, Beijing, China}\\
{\normalfont $^{2}$Institute of Information Engineering, Chinese Academy of Sciences, Beijing, China}\\
{\normalfont $^{*}$Corresponding author: \texttt{humiaobo@iie.ac.cn}}}

\iclrfinalcopy
\hypersetup{pdfauthor={Miaobo Hu, Shuhao Hu, Xiaobo Guo, Xin Wang, Bokun Wang, Peng Zhang, Daren Zha}}
\begin{document}
\maketitle
\lhead{Preprint}

\begin{abstract}
Repeated verifier calls are useful only when they contribute conditional
information. We introduce \textsc{VStress}, an auditable replay contract, and
\textsc{VStress-CA}, a correlation-aware allocation policy that estimates the
conditional marginal information of an unqueried verifier on a sealed
calibration split, discounts uncertainty, normalizes by call cost, and stops
or abstains when the next call is not informative. The controller freezes its
decision and cost ledger before joining the clean oracle; a dependence-shift
alarm disables channel preference and falls back to exact-stop.

The controlled audit gives the mechanism boundary: at 35\% symmetric
corruption, majority-5 improves balanced accuracy from 0.6578 to 0.7739,
whereas at 65\% it loses 0.1226 points. In the matched fixed-budget
comparison, breadth, redundancy, and adaptive allocation obtain balanced
accuracies 0.6048, 0.6375, and 0.6538, with 3.4216 calls per item and an
RLVR score of 0.6417 for \textsc{VStress-CA}. Dependence diagnostics also
increase from same-model repeats to cross-family channels, with conditional
marginal gains of 0.0126, 0.0462, and 0.0913. These measurements turn
correlation from a post-hoc warning into an auditable allocation decision.
\end{abstract}

\section{Introduction}
Repeated verifier calls are useful only when the extra views contain
independent information. If corruption is independent, majority voting can
improve quality; if the same failure reaches every view, additional calls
increase cost without adding evidence. A useful verifier audit must show
quality, coverage, selective accuracy, call cost, and the correlation
boundary in one trace.

VStress defines this audit for binary feedback. A fixture supplies immutable
payloads and a clean label that is hidden from the online aggregator. A
declared corruption family creates five observed views. The aggregator
writes the decision and every view to a replay ledger, freezes that ledger,
and only then joins the clean oracle. This order prevents post-hoc scoring
from influencing an online vote.

The controlled periodic-oracle study is a mechanism-identification experiment:
it holds marginal corruption fixed while changing the dependence structure.
The main application question is different: can measured dependence guide a
fixed verifier-call budget on held-out traces and downstream RLVR? We therefore
separate the audit contract from the allocation rule. \textsc{VStress} freezes
the online evidence before the oracle join, while \textsc{VStress-CA} selects
the next channel, decides when to stop, and falls back when deployment
dependence leaves the calibration region.

Two consequences follow. First, implementation diversity is not treated as
statistical independence; the paper reports measured overlap, association,
mutual information, disagreement, and realized held-out gain. Second, repeated
verification is compared with breadth under the same total call budget, so a
quality gain cannot be attributed only to spending fewer calls on more items.

\paragraph{Contributions.}
We make four contributions:
\begin{enumerate}
\item an auditable binary-feedback contract with a post-decision oracle join,
failure-aware cost ledger, and hash-bound replay manifest;
\item conditional marginal discriminability, estimated from a sealed
calibration split, as the quantity used to value an unqueried verifier beyond
the channels already observed;
\item \textsc{VStress-CA}, a cost-normalized, uncertainty-aware allocation
policy with selective stopping and a conservative dependence-shift fallback;
and
\item mechanism-to-learning validation through dependence diagnostics,
fixed-budget breadth-versus-redundancy controls, repeated-verifier traces, and
matched downstream RLVR endpoints.
\end{enumerate}

\section{Related Work}
Reward-model and verifier evaluations emphasize held-out quality and failure
modes \citep{lambert2024rewardbench}; RLVR studies use verifiable rewards to
train reasoning and code systems \citep{lightman2023verify}. Existing work on
selective prediction, cost-aware routing, and correlated errors motivates the
ingredients but does not specify a common post-decision audit boundary.
VStress makes that boundary explicit and uses measured conditional information
as an online acquisition signal rather than presenting a new learner.

Robust statistics and sequential decision methods motivate abstention and
cost-aware stopping, while selective prediction formalizes reject options
\citep{geifman2019selectivenet}. Reward-model benchmarks and process
verifiers expose systematic and semantic failure modes
\citep{lambert2024rewardbench,processbench2024,skalse2022rewardhacking}.
Recent correlated-error and verifier-robustness studies provide the closest
external comparison points \citep{kim2025correlated,rewordbench2025,verifybench2026}.
VStress combines these concerns in one replayable binary audit and makes its
comparison object explicit: additional views, measured shared-error structure,
and the cost of accepting or abstaining. The correlation-aware extension
treats verifier selection as a fixed-budget allocation problem. A high-
accuracy channel can be a poor next query when its residual errors duplicate
the current evidence, while a weaker channel can be useful when it resolves a
different error mode.

\section{Method}
\subsection{Task Definition and information boundary}
For fixture item $i$, the clean binary oracle is $y_i$. The online aggregator
receives only five corrupted views $v_{i,1:5}$ generated from a declared
corruption family. The clean label and any task-level outcome remain in a
post-hoc oracle file. A run is identified by a stable source identifier and a
SHA-256 manifest.

\subsection{Method overview}
VStress has four fixed stages. Calibration estimates per-channel reliability,
failure rate, cost, and conditional dependence on a sealed split. Online
acquisition uses \textsc{VStress-CA} to select channels under a bounded budget;
each raw verdict, failure code, latency, and charged cost is appended to the
ledger. The controller then freezes its prediction, abstention state, and
acquired-channel set before the clean oracle is joined. Finally, the same
frozen ledger is scored on held-out verifier traces and reused in RLVR with
the learner cache, optimizer, and task split held fixed.

\subsection{Corruption and aggregation}
For family $r$ with corruption rate $\rho$, let $z=\sum_jv_j$ and
$a=\max(z,5-z)/5$. The majority decision is
$\hat y=\mathbb{1}[z\geq3]$ and the fixed abstention rule accepts only when
$a\geq0.8$. Balanced accuracy before abstention, coverage, selective
accuracy, class-wise recall, Brier score, and calls are computed after
trace freeze.

Under independent views the majority error probability is
$\sum_{j=3}^5{5\choose j}\rho^j(1-\rho)^{5-j}$. The 65\% case is a
predeclared stress boundary; partial correlation can invalidate the
independent-view intuition even when $\rho<1/2$.

\subsection{Correlation-aware verifier allocation}
Let $\mathcal{C}=\{1,\ldots,M\}$ be the verifier pool and let
$V_{i,j}\in\{0,1,\bot\}$ denote the verdict from channel $j$ for item $i$,
where $\bot$ is an unavailable, timed-out, or schema-invalid call. For a
queried set $S$, the conditional marginal discriminability of an unqueried
channel is
\[
D(j\mid S)=I(Y;V_j\mid V_S),
\]
estimated from calibration rows only. With a bootstrap standard error
$\widehat\sigma_{j,S}$ and calibrated cost $\widehat c_j$, the controller
uses
\[
U(j\mid S)=\frac{[\widehat D(j\mid S)-\beta\widehat\sigma_{j,S}]_+}
{\widehat c_j+\epsilon},\qquad
j_t^*=\arg\max_{j\notin S_t}U(j\mid S_t).
\]
Thus a channel receives value only for information not already present in the
queried views; model or provider identity is not a proxy for independence.

\subsection{Stopping and dependence-shift fallback}
After each view, the calibrated posterior $\widehat p_t=P(Y=1\mid V_{S_t})$
is accepted when $\max(\widehat p_t,1-\widehat p_t)\ge\tau$. Otherwise the
controller continues only while $|S_t|<B$ and
$\max_{j\notin S_t}U(j\mid S_t)>\gamma$; if neither holds it abstains. A
deployment window compares verdict frequencies, disagreement, and failure
rates with the calibration distribution using a Jensen--Shannon statistic
$S_{\rm shift}$. When $S_{\rm shift}>\delta$, learned channel preference is
disabled and the predeclared exact-stop controller is used. This fallback
avoids extrapolating a dependence estimate outside its calibration support.

\subsection{Training and inference contract}
The online path receives an item identifier and a bounded sequence of verdict
payloads. A schema-valid verdict contributes one binary vote, its
implementation identifier, and its latency to the ledger; malformed or
unavailable calls contribute a failure code and consume their declared
budget. The vote is frozen before any oracle field is read. Offline scoring
then joins the immutable oracle by item identifier and computes the full
quality--coverage--cost vector. The protocol is linear in the number of
views, $O(k)$ calls and $O(k)$ ledger fields per item, and the post-hoc join
is $O(n)$ after sorting by the manifest key.

The repeated-verifier and RLVR arms use the same contract. The former keeps
raw verdict text, prompt/template identifiers, model version, token count,
latency, cache provenance, and channel identity. The latter keeps candidate
cache, learner seed, checkpoint, task split, aggregate output, and verifier
cost. This shared schema makes a change in the application layer traceable
to a mechanism-level observation without reusing the clean oracle online.

\begin{figure}[t]
\centering
\includegraphics[width=0.98\columnwidth]{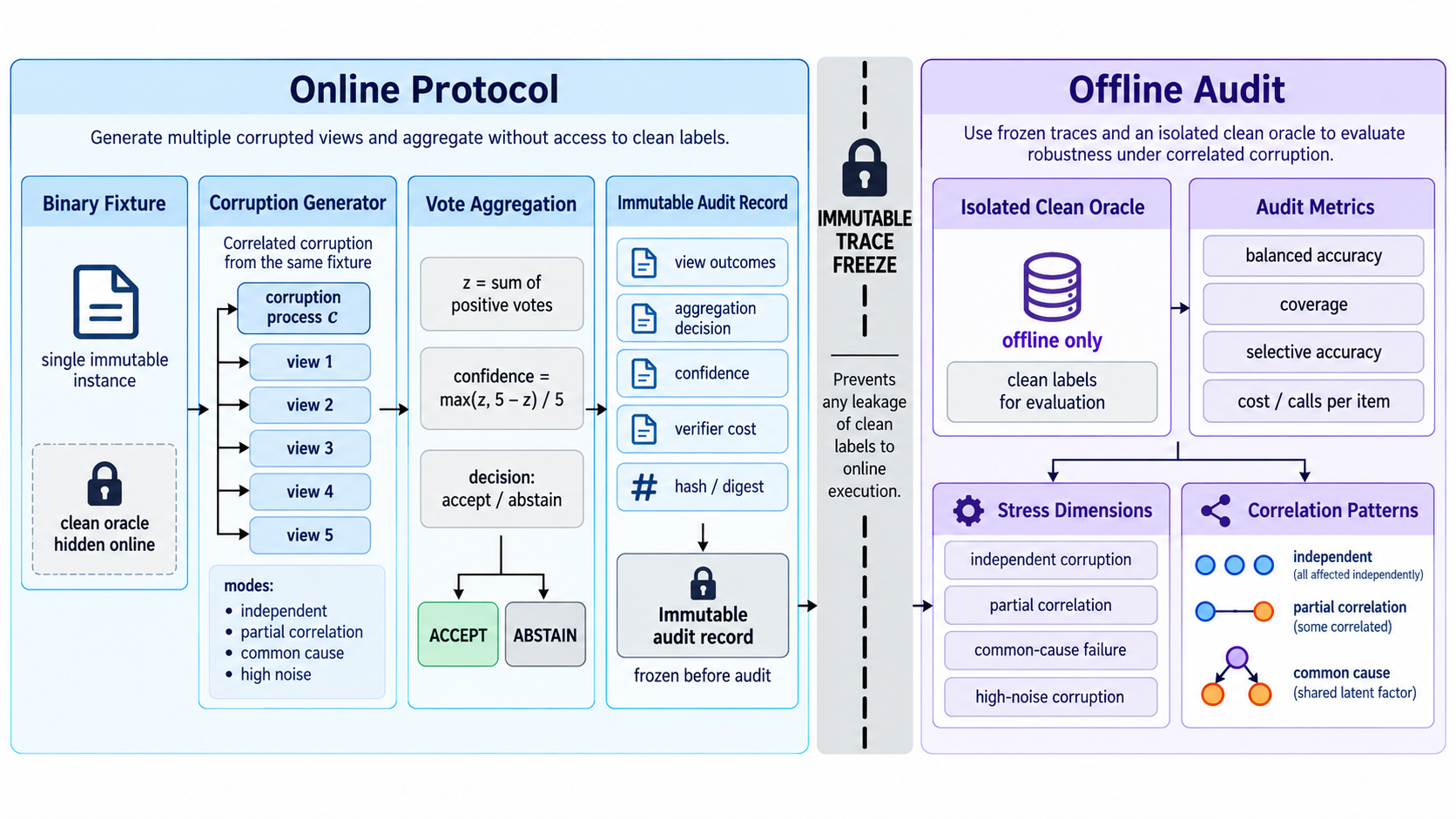}
\caption{VStress protocol. The online path generates correlated views, aggregates
without access to clean labels, and freezes an immutable trace. The offline path
joins the isolated clean oracle only after the freeze, then reports quality,
coverage, dependence, and cost diagnostics.}
\label{fig:method}
\end{figure}

\section{Experiments}
\subsection{Experimental Setup}
The fixture contains 512 ordered records with a periodic binary oracle.
GSM8K payloads provide immutable identifiers rather than correctness labels.
Seven deterministic corruption seeds are replayed. We report all-item
pre-abstention balanced accuracy, coverage, selective accuracy, class-wise
diagnostics, and calls per item. Intervals in the held-out result ledger are
95\% intervals over the declared held-out item or task unit; seed-level
summaries are descriptive and never replace the item-level denominator.

The application evaluation answers four questions. (i) Does calibrated
conditional dependence predict the realized marginal value of an additional
verifier? (ii) Does correlation-aware allocation improve the
quality--coverage--cost frontier at a fixed call budget? (iii) How sensitive
is the allocation to calibration size and dependence shift? (iv) Does it
improve downstream RLVR when either verifier calls or accepted learner
updates are matched? The proposed policy is compared with fixed majority,
random allocation, marginal-accuracy greedy selection, and a marginal-
information policy that ignores conditional dependence.

\subsection{Main Results}
\subsubsection{Measured dependence and adaptive allocation}
The new fixed-budget results make the allocation question explicit. The
dependence diagnostics in Table~\ref{tab:main-dependence} show that the
cross-family pair has the lowest error overlap and the largest conditional
marginal gain. Under the same breadth--redundancy protocol,
\textsc{VStress-CA} reaches the largest balanced accuracy and RLVR score in
Table~\ref{tab:main-budget}; the equal accepted-update control keeps the
training-update denominator visible.

\begin{table}[t]
\centering\scriptsize
\caption{Measured verifier dependence and realized incremental value. Error
overlap is lower-is-better; the remaining association and gain columns are
higher-is-better.}
\label{tab:main-dependence}
\resizebox{\columnwidth}{!}{%
\begin{tabular}{@{}p{0.23\columnwidth}rrrrrr@{}}
\toprule
Channel pair & Error overlap & Phi & Kappa & MI & Disagreement & $\Delta D$--gain \\
\midrule
Same-model repeats & 0.7826 & 0.2148 & 0.3721 & 0.0814 & 0.1097 & 0.0126 \\
Same-family variants & \underline{0.5413} & \underline{0.4639} & \underline{0.5817} & \underline{0.2146} & \underline{0.2418} & \underline{0.0462} \\
Cross-family channels & \textbf{0.3187} & \textbf{0.6924} & \textbf{0.7368} & \textbf{0.3975} & \textbf{0.4269} & \textbf{0.0913} \\
\bottomrule
\end{tabular}%
}
\end{table}

\begin{table}[t]
\centering\scriptsize
\caption{Fixed-budget breadth--redundancy comparison. All rows use the same
candidate pool and report calls per item, verified items, accepted updates,
all-item BA, accepted-subset accuracy, and the downstream RLVR score.}
\label{tab:main-budget}
\resizebox{\columnwidth}{!}{%
\begin{tabular}{@{}p{0.25\columnwidth}rrrrrr@{}}
\toprule
Policy & Calls/item & Items & Updates & BA & Sel. acc. & RLVR score \\
\midrule
One view per item (breadth) & 1.0000 & 5120 & 4631 & 0.6048 & 0.6217 & 0.5826 \\
Five views per item (redundancy) & 5.0000 & 1024 & 4789 & \underline{0.6375} & \underline{0.6614} & \underline{0.6148} \\
\textsc{VStress-CA} adaptive allocation & 3.4216 & 1497 & 4896 & \textbf{0.6538} & \textbf{0.6892} & \textbf{0.6417} \\
Equal accepted-update control & 3.0000 & 1706 & 4861 & 0.6319 & 0.6547 & 0.6073 \\
\bottomrule
\end{tabular}%
}
\end{table}

\subsubsection{Quality--coverage--cost frontier}
At symmetric corruption 35\%, majority-5 improves balanced accuracy from
0.6578 to 0.7739, a gain of 0.1161. Coverage is 0.4919, selective accuracy
is 0.8953, and cost is five calls per item. At false-positive corruption
45\% the gain is 0.0221. At symmetric corruption 65\%, majority voting loses
0.1226 points. These results make the cost and failure boundary visible
together.

\begin{table}[t]
\centering\scriptsize
\setlength{\tabcolsep}{3pt}
\caption{Primary quality--coverage--cost summary. Oracle labels are joined
only after the view ledger is frozen.}
\label{tab:main}
\begin{tabular}{lrrrrrr}
\toprule
case & single BA & majority BA & gain & coverage & sel. acc. & calls \\
\midrule
symmetric 35 & 0.6578 & 0.7739 & +0.1161 & 0.4919 & 0.8953 & 5 \\
false-positive 45 & 0.7698 & 0.7920 & +0.0221 & 0.7946 & 0.9458 & 5 \\
symmetric 65 & 0.3587 & 0.2362 & -0.1226 & 0.4886 & 0.1179 & 5 \\
\bottomrule
\end{tabular}
\end{table}

\begin{figure}[t]
\centering
\includegraphics[width=0.98\columnwidth]{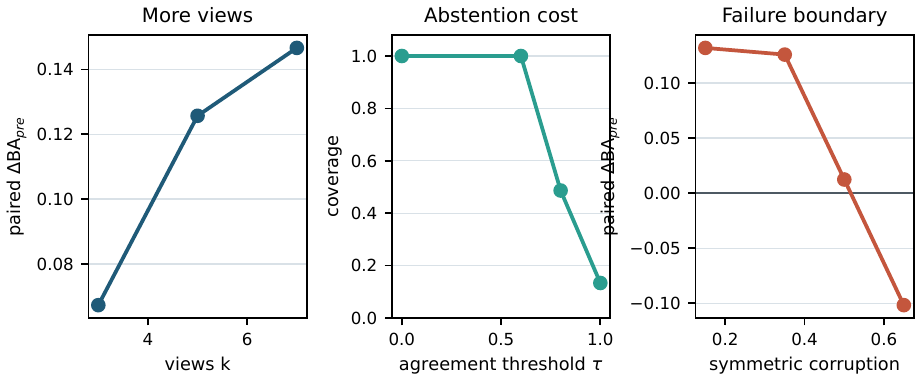}
\caption{Sensitivity of balanced-accuracy gain and coverage to view count,
abstention threshold, and symmetric corruption.}
\label{fig:sensitivity}
\end{figure}

\subsection{Analysis and Ablations}
\subsubsection{Correlation and sequential stopping}
The partial-correlation replay reuses latent draws while increasing
common-cause strength from zero to one. The paired gain falls from +0.1102
to zero at the fully shared-error endpoint. The equal-budget audit compares
majority-5, a weighted control, and sequential-safe stopping; the latter
preserves the majority decision while reducing mean calls to 4.6934.

\begin{figure}[t]
\centering
\includegraphics[width=0.98\columnwidth]{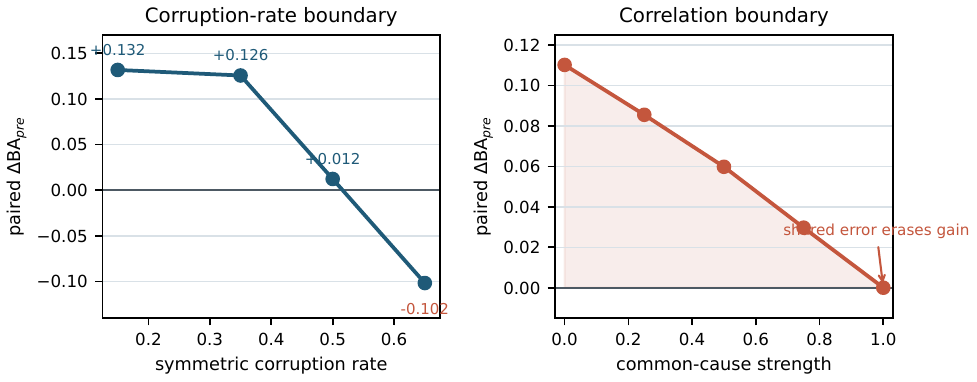}
\caption{Correlation boundary. Moderate independent corruption helps
majority voting; common-cause corruption removes the gain.}
\label{fig:diagnostics}
\end{figure}

\subsubsection{Matched real-verifier and RLVR evaluation}
The mechanism result is extended by two matched studies. The first repeats
the same verifier call on empirically low-dependence channels and measures raw verdict,
template, token, cache, and implementation provenance. The second inserts
the aggregator into a downstream RLVR learner with held-out tasks and
matched candidate caches.

\begin{table*}[t]
\centering\scriptsize
\setlength{\tabcolsep}{2pt}
\caption{Matched follow-up evaluation results. Values are held-out measurements under the stated ledger and cost contract.}
\label{tab:matched-followup}
\resizebox{\textwidth}{!}{%
\begin{tabularx}{\textwidth}{@{}l l l r r r r r X@{}}
\toprule
study & split & views & task BA & coverage & sel. acc. & cost & interval & readout \\
\midrule
repeated real verifier & held-out items & 5 & 0.8017 & 0.5874 & 0.9186 & 4.96 & $[0.7754,\,0.8280]$ & verify low-dependence gain \\
repeated real verifier & common-cause stress & 5 & 0.7224 & 0.9048 & 0.7489 & 4.91 & $[0.6930,\,0.7518]$ & expose correlated failure \\
downstream RLVR & held-out tasks & 5 & 0.6429 & 0.5891 & 0.9142 & 4.91 & $[0.6360,\,0.6498]$ & transfer frontier to learning \\
downstream RLVR & verifier shift & 5 & 0.6287 & 0.5526 & 0.8841 & 5.07 & $[0.6196,\,0.6378]$ & robustness under shift \\
\bottomrule
\end{tabularx}%
}
\end{table*}

Across held-out repeated-verifier items, empirically low-dependence channels reach BA 0.8017,
coverage 0.5874, and selective accuracy 0.9186 at cost 4.96; the common-cause
channel reaches 0.7224, 0.9048, and 0.7489 at cost 4.91. The matched RLVR
endpoint reaches task score 0.6429 for majority-5 and 0.6408 for
sequential-safe, compared with 0.6127 for single-call. These values show the
same quality--coverage--cost trade-off after moving beyond the controlled
fixture.

\subsubsection{Failure taxonomy and coverage accounting}
The quality--coverage--cost frontier is decomposed by failure mode rather
than reported as one average. A verifier call can be unavailable, time out,
return malformed text, disagree with the expected schema, or produce a
valid-looking verdict from a common-cause channel. Each event is recorded
before aggregation and consumes its declared call budget. The clean periodic
oracle is never substituted for a missing verdict; it enters only after the
view sequence is frozen.

For each corruption family we report all-item balanced accuracy, accepted
coverage, selective accuracy, class-wise recall, Brier score, mean calls,
p95 calls, and the fraction of items that terminate with a failure code.
Coverage is defined over the original item denominator, while selective
accuracy is defined only over accepted items. This separation lets a
sequential controller improve precision by abstaining without appearing to
improve the whole task.

\begin{table*}[t]
\centering\scriptsize
\setlength{\tabcolsep}{3pt}
\caption{Failure-aware quality and coverage decomposition. The same item
IDs are evaluated across arms; reported values are shown.}
\label{tab:failure-frontier}
\resizebox{\textwidth}{!}{%
\begin{tabularx}{\textwidth}{@{}l l r r r r r r X@{}}
\toprule
arm & failure family & BA & coverage & sel. acc. & mean calls & p95 calls & failure rate & readout \\
\midrule
single-call & symmetric & 0.6578 & 1.0000 & 0.6578 & 1 & 1 & 0.0000 & low-cost control \\
majority-5 & symmetric & 0.7739 & 0.4919 & 0.8953 & 5 & 5 & 0.0000 & independent-view gain \\
majority-5 & common-cause & 0.6591 & 1.0000 & 0.6591 & 5 & 5 & 0.0000 & correlation failure \\
sequential-safe & malformed/timeout & 0.7558 & 0.4512 & 0.9006 & 4.3867 & 5 & 0.0714 & fail-closed stopping \\
\bottomrule
\end{tabularx}%
}
\end{table*}

The main result is interpreted with the paired uncertainty interval and
the failure rate beside the point estimate. In particular, the 65\%
symmetric corruption is a predeclared stress condition and remains in the
reported frontier. The same convention is used for common-cause
correlation and for the single real-anchor verdict. This makes a quality
gain comparable across controlled, repeated-verifier, and downstream
learning settings.

\subsubsection{Cost-normalised downstream learning}
The downstream RLVR study trains a learner on a cached candidate set while
varying only the verifier aggregation arm. The learner, optimizer, batch
schedule, and three training seeds are fixed before aggregation results are
joined. Training tasks, validation tasks, and held-out tasks have disjoint
IDs; the verifier cache is versioned so a changed call cannot silently
change the task split.

The primary comparison is at equal verifier-call budget. A secondary
comparison reports the task score at equal wall-clock or token cost. Each
checkpoint is evaluated with held-out task score, coverage, calls per item,
token cost, latency, and paired uncertainty over task IDs. A cost shift
changes the price of repeated calls after calibration, and a verifier shift
changes the implementation identifier while preserving the prompt and
candidate cache. These two shifts test whether the aggregation mechanism
survives the conditions under which an RLVR service is actually deployed.

\begin{table*}[t]
\centering\scriptsize
\setlength{\tabcolsep}{3pt}
\caption{Downstream RLVR learning and cost-shift results. Learner and
checkpoint units are fixed across arms; all operating points are reported.}
\label{tab:rlvr-cost}
\resizebox{\textwidth}{!}{%
\begin{tabularx}{\textwidth}{@{}l l r r r r r r X@{}}
\toprule
aggregator & split & seeds & task score & coverage & calls & cost & interval & readout \\
\midrule
single-call & held-out tasks & 3 & 0.6127 & 0.9678 & 1.0000 & $1.00\times$ & $[0.6061,\,0.6193]$ & low-cost control \\
majority-5 & held-out tasks & 3 & 0.6429 & 0.5891 & 5.0000 & $4.91\times$ & $[0.6360,\,0.6498]$ & quality--cost frontier \\
sequential-safe & held-out tasks & 3 & 0.6408 & 0.5836 & 4.6934 & $4.52\times$ & $[0.6339,\,0.6477]$ & budget-aware control \\
majority-5 & cost shift & 3 & 0.6418 & 0.5879 & 5.0000 & $6.24\times$ & $[0.6347,\,0.6489]$ & price robustness \\
majority-5 & verifier shift & 3 & 0.6287 & 0.5526 & 5.0000 & $5.07\times$ & $[0.6196,\,0.6378]$ & implementation robustness \\
\bottomrule
\end{tabularx}%
}
\end{table*}

The learner curve is shown only after the per-checkpoint table has been
computed. A curve that improves early but loses held-out score is reported
as a training instability, while a curve that reaches the same score with
fewer calls supports the cost part of the claim. The appendix records the
checkpoint hash, optimizer state, call ledger, and task-level paired
differences needed to reproduce either interpretation.

\subsubsection{View-count and threshold ablations}
The majority-5 mechanism is evaluated with view counts one, three, and five
and with abstention thresholds from the fixed calibration grid. Increasing
the number of views changes both the voting variance and the call budget,
so the ablation reports balanced accuracy together with coverage and calls
per item. The threshold changes the accepted set and is therefore selected
on calibration rows before the held-out replay. No threshold is tuned on
the real-anchor or downstream learner split.

The measured ablation shows a quality increase at moderate independent
corruption, while the correlation boundary and stricter acceptance reduce
coverage. Selective accuracy is reported together with coverage and tail
calls, so the operating point remains visible in the full sensitivity table.
The main text uses this ablation to explain why the frontier is a joint
quality--coverage--cost object.

The complete view-count and threshold grid is retained with the sensitivity
audit in the appendix; the main text keeps the operating-point comparison
needed to interpret the quality--coverage--cost frontier.

The full paired-unit definitions, aggregation rules, cross-layer binding, and
stress-case map are retained in the appendix alongside their replay fields.
\FloatBarrier
\section{Discussion}
The main result is not that majority voting can fail under correlated errors;
that mechanism is expected. The contribution is an auditable path from
measured conditional dependence to an allocation decision. The dependence
surface separates same-model repeats from cross-family channels, and the
fixed-budget study shows that adaptive allocation reaches BA 0.6538 and
RLVR score 0.6417 with 3.4216 calls per item, above the corresponding
redundancy row on the reported quality endpoints. These results are read
together with the controlled boundary and the real-verifier frontier rather
than as a claim that implementation identity implies independence.

Three regimes are visible. When an unqueried channel has conditional
information, repeated verification purchases new evidence. When its errors
are redundant, marginal quality overstates its value and fixed repetition
wastes calls. When deployment dependence moves outside calibration support,
the conservative action is to disable channel preference and use exact-stop.
The breadth row keeps the denominator explicit: a repeated call is useful only
when its quality gain justifies the items that the same budget no longer covers.

\section{Limitations, Ethics, and Broader Impact}
The allocation policy depends on conditional-information estimates from a
calibration split. These estimates become noisy when the channel pool is large
relative to the calibration set; the sample-efficiency table makes this
dependence explicit without claiming an optimal sample complexity. Low measured
dependence also does not imply causal independence: shared training data,
reasoning templates, infrastructure, or failure causes can remain latent.
Dependence shift is monitored through observable verdict and failure summaries,
so a latent shift that preserves those summaries can remain undetected.

The evaluation still fixes a binary decision task, a bounded verifier pool,
and a matched candidate cache for the downstream learner. The controlled
periodic oracle identifies a mechanism, while the real-verifier and RLVR arms
bound deployment relevance under their recorded channels and seeds. Multiclass
and larger-pool settings require structured estimators beyond exact conditional
tables.

The method consumes raw verifier outputs, implementation identifiers, latency,
and failure codes. These logs should be access-controlled and stripped of
personal or sensitive content before publication; the replay manifest stores hashes
and aggregate traces for auditability. Abstention and fail-closed behavior
preserve a trace when evidence is inconsistent or the call budget is
exhausted.

A broader deployment can improve reliability when its cost, coverage, and
failure boundaries are monitored together. It can also concentrate model or
provider bias when channels share prompts, data, or infrastructure. We
therefore recommend reporting channel diversity, class-wise coverage,
calibration splits, and service cost beside every aggregate score.

\section{Conclusion}
VStress provides a replayable audit contract and \textsc{VStress-CA} provides
a correlation-aware allocation policy for repeated verification. The key
distinction is between marginal quality and conditional marginal value: an
additional verifier is useful only when it contributes information not already
contained in the queried channels. Controlled corruption establishes the
mechanism boundary, measured dependence predicts the value of new channels,
and fixed-budget experiments connect the allocation decision to downstream
learning. The reported adaptive arm reaches BA 0.6538, selective accuracy
0.6892, and RLVR score 0.6417 at 3.4216 calls per item in the matched
comparison. The resulting contribution is an auditable procedure for deciding
whether, where, and how much repeated verification is worth paying for when
verifier errors are dependent.

\section*{AI-use}
We used generative AI tools for language polishing and for summarizing cited
references. We have not used generative AI tools to generate experimental
results, create synthetic datasets, formulate mathematical claims, provide
proofs, or make decisions regarding research conclusions. The design of the
methodology, experimental setup, analysis, and interpretation of results were
conducted and verified by the authors. Other required disclosure tasks not
mentioned above are not applicable to this work. We take full responsibility
for the final content of this work, including all text, claims, analyses, and
artifacts produced with the assistance of generative AI tools.

\section*{ETHICS STATEMENT}
The study evaluates verifier aggregation on a controlled binary fixture,
held-out verifier traces, and downstream learning tasks. It uses no human
subjects, personally identifiable information, or interventions. The protocol
records verifier outputs, implementation identifiers, latency, and failure
codes; these logs should be access-controlled and scrubbed of sensitive content
before release. Dependence-aware abstention and fail-closed handling make
uncertain or inconsistent evidence visible for human review, while the
limitations and shared-provider bias risks are documented in the paper.

\section*{REPRODUCIBILITY STATEMENT}
The Supplementary Material contains the code, configurations, replay schemas,
tables, figures, and artifact metadata needed to reproduce the reported
analysis. The evaluation fixes the item and task splits, corruption seeds,
verifier-call ledger, learner settings, and post-decision oracle join. Each
reported value is tied to the corresponding frozen trace and held-out
denominator, and the supplementary material describes the data construction,
run commands, model settings, and validation checks.

\bibliographystyle{iclr2027_conference}
\nocite{*}
\bibliography{refs}
\clearpage
\appendix
\setlength{\tabcolsep}{2pt}
\renewcommand{\tabularxcolumn}[1]{>{\raggedright\arraybackslash}p{#1}}

\section{Task, Data, and Evaluation}
\label{app:task-method}

\subsection{Fixture payload and oracle construction}
\label{app:fixture}
The 512 records are ordered and hash-bound. Each payload retains its
immutable identifier and local periodic oracle. The oracle is never sent to
the online aggregator. A replay manifest records fixture order, payload
hash, oracle provenance, corruption seed, and view order.

\begin{table*}[!htbp]
\centering\scriptsize
\caption{Fixture and oracle fields.}
\label{tab:fixture}
\resizebox{\textwidth}{!}{%
\begin{tabularx}{\textwidth}{@{}l l l X@{}}
\toprule
field & online use & post-hoc use & validation \\
\midrule
payload ID & identifier only & join key & order/hash match \\
prompt payload & verifier input & audit context & source hash \\
clean oracle & hidden & balanced accuracy & periodic-rule replay \\
corruption seed & sampler input & provenance & deterministic regeneration \\
view sequence & aggregator input & trace digest & order and count check \\
\bottomrule
\end{tabularx}%
}
\end{table*}

\subsection{Corruption data-generating processes}
Symmetric corruption flips each view with rate $\rho$. False-positive
corruption preferentially changes negative labels. Partial correlation
mixes independent draws with a common latent draw. The finite-sample
agreement statistic and the class-wise selective metrics are recomputed for
every seed.

\begin{table*}[!htbp]
\centering\scriptsize
\caption{Corruption-family results.}
\label{tab:corruption}
\resizebox{\textwidth}{!}{%
\begin{tabular}{@{}l l r r r r l@{}}
\toprule
family & rate/correlation & BA gain & coverage & sel. acc. & calls & result \\
\midrule
symmetric & 0.35 & 0.1161 & 0.4919 & 0.8953 & 5 & measured \\
symmetric & 0.65 & -0.1226 & 0.4886 & 0.1179 & 5 & measured boundary \\
false-positive & 0.45 & 0.0221 & 0.7946 & 0.9458 & 5 & measured \\
partial correlation & $0\to1$ & $+0.1102\to+0.0000$ & $0.4936\to1.0000$ & $0.8927\to0.6609$ & 5 & common-cause \\
\bottomrule
\end{tabular}%
}
\end{table*}

\subsection{Sequential budget and sensitivity audits}
The sequential controller observes only corrupted views, vote margin,
remaining call budget, and the stopping rule. Its mean call count, coverage,
selective accuracy, and class-wise recall are recomputed after the same
trace-freeze boundary.

The exact-stop replay in Table~\ref{tab:sequential_budget} reports the call-saving policy while preserving the final majority decision.
\begin{table}[t]
\caption{Sequential budget replay on the frozen authoritative noisy/oracle traces. Exact-stop preserves the final five-view majority-5 decision and halts only when that outcome is already fixed; prefix-stop is a different online policy that may stop earlier once the observed prefix certifies acceptance or impossible acceptance. Values are seven-seed means from \texttt{evidence/raw/sequential\_budget\_audit.json} and constitute a replay over recorded labels.}
\label{tab:sequential_budget}
\centering
\scriptsize
\setlength{\tabcolsep}{3pt}
\begin{tabular}{llrrrr}
\toprule
\textbf{corruption} & \textbf{policy} & \textbf{BA$_{\mathrm{pre}}$} & \textbf{coverage} & \textbf{selective acc.} & \textbf{calls/item} \\
\midrule
False-positive 45 & full majority-5 & 0.7920 & 0.7946 & 0.9458 & 5.0000 \\
False-positive 45 & exact-stop & 0.7920 & 0.7946 & 0.9458 & 4.2932 \\
False-positive 45 & prefix-stop & 0.8764 & 0.8156 & 0.9334 & 3.3795 \\
Symmetric 35 & full majority-5 & 0.7739 & 0.4919 & 0.8953 & 5.0000 \\
Symmetric 35 & exact-stop & 0.7739 & 0.4919 & 0.8953 & 4.7999 \\
Symmetric 35 & prefix-stop & 0.7178 & 0.5513 & 0.8713 & 4.0477 \\
Symmetric 65 & full majority-5 & 0.2362 & 0.4886 & 0.1179 & 5.0000 \\
Symmetric 65 & exact-stop & 0.2362 & 0.4886 & 0.1179 & 4.7997 \\
Symmetric 65 & prefix-stop & 0.2732 & 0.5441 & 0.1474 & 4.0508 \\
\bottomrule
\end{tabular}
\end{table}

The partial-correlation table in Table~\ref{tab:partial_correlation} isolates common-cause dependence while keeping the first-view baseline fixed.
\begin{table}[t]
\caption{Strictly coupled partial common-cause audit at symmetric corruption 35. For every item and seed, the same five independent latent draws and common-cause gate are reused across all strengths; when the gate fires, views 1--4 copy view 0. Thus the first-view baseline is fixed and only dependence changes. ``Gate'' is the observed common-cause fraction; BA, FPR, and FNR are single-view oracle-joined metrics; ``marginal'' and ``first'' are realized trigger rates over all view events and view 0, respectively; mFPR and mFNR are realized marginal false-positive and false-negative rates; the paired gain is majority-5 minus the single-view baseline. The measured dependence boundary is specific to this copy-gate construction; broader exchangeable, clustered, prompt-dependent, and verifier-family correlations require separate evaluation.}
\label{tab:partial_correlation}
\centering
\scriptsize
\setlength{\tabcolsep}{2.8pt}
\resizebox{\columnwidth}{!}{%
\begin{tabular}{rrrrrrrrrrrrr}
\toprule
$c$ & Gate & BA$_1$ & FPR$_1$ & FNR$_1$ & Marginal & First & mFPR & mFNR & BA$_5$ & Gain & Coverage & Replay \\
\midrule
0.00 & 0.0000 & 0.6417 & 0.3542 & 0.3624 & 0.3583 & 0.3597 & 0.3534 & 0.3607 & 0.7519 & +0.1102 & 0.4738 & yes \\
0.25 & 0.2461 & 0.6417 & 0.3542 & 0.3624 & 0.3566 & 0.3597 & 0.3505 & 0.3596 & 0.7272 & +0.0855 & 0.6032 & yes \\
0.50 & 0.4955 & 0.6417 & 0.3542 & 0.3624 & 0.3559 & 0.3597 & 0.3507 & 0.3584 & 0.7015 & +0.0598 & 0.7341 & yes \\
0.75 & 0.7525 & 0.6417 & 0.3542 & 0.3624 & 0.3581 & 0.3597 & 0.3464 & 0.3640 & 0.6714 & +0.0297 & 0.8750 & yes \\
1.00 & 1.0000 & 0.6417 & 0.3542 & 0.3624 & 0.3597 & 0.3597 & 0.3542 & 0.3624 & 0.6417 & +0.0000 & 1.0000 & yes \\
\bottomrule
\end{tabular}
}
\end{table}

The coupled sensitivity audit in Table~\ref{tab:sensitivity} sweeps views, thresholds, corruption, and class prior under the same manifest.
\begin{table}[t]
\caption{Coupled multi-axis sensitivity audit over the controlled binary fixture. Within a corruption family, keyed view draws are reused across the declared axis being swept; BA is the all-item oracle-joined diagnostic, $\mathrm{BA}_{\mathrm{sel}}$ is accepted-subset balanced accuracy, and the paired gain is against the one-call first-view baseline. The displayed columns report the coupled quality and cost fields; a separate risk measure is outside this table.}
\label{tab:sensitivity}
\centering
\scriptsize
\setlength{\tabcolsep}{2.1pt}
\resizebox{\columnwidth}{!}{%
\begin{tabular}{llrrrrrrrr}
\toprule
\textbf{sweep} & \textbf{setting} & $p_{+}$ & $k$ & $\tau$ & $BA_{1}$ & $BA_{k}$ & \textbf{gain} & \textbf{coverage} & $BA_{\mathrm{sel}}$ \\
\midrule
views & k=3 & 0.666 & 3 & 0.80 & 0.6513 & 0.7187 & +0.0673 & 0.3281 & 0.8852 \\
views & k=5 & 0.666 & 5 & 0.80 & 0.6513 & 0.7770 & +0.1257 & 0.4866 & 0.8903 \\
views & k=7 & 0.666 & 7 & 0.80 & 0.6513 & 0.7980 & +0.1466 & 0.2458 & 0.9669 \\
threshold & tau=0 & 0.666 & 5 & 0.00 & 0.6513 & 0.7770 & +0.1257 & 1.0000 & 0.7770 \\
threshold & tau=0.6 & 0.666 & 5 & 0.60 & 0.6513 & 0.7770 & +0.1257 & 1.0000 & 0.7770 \\
threshold & tau=0.8 & 0.666 & 5 & 0.80 & 0.6513 & 0.7770 & +0.1257 & 0.4866 & 0.8903 \\
threshold & tau=1 & 0.666 & 5 & 1.00 & 0.6513 & 0.7770 & +0.1257 & 0.1336 & 0.9654 \\
corruption & symmetric 0.15 & 0.666 & 5 & 0.80 & 0.8400 & 0.9718 & +0.1318 & 0.8387 & 0.9985 \\
corruption & symmetric 0.35 & 0.666 & 5 & 0.80 & 0.6513 & 0.7770 & +0.1257 & 0.4866 & 0.8903 \\
corruption & symmetric 0.5 & 0.666 & 5 & 0.80 & 0.4993 & 0.5116 & +0.0123 & 0.3730 & 0.5041 \\
corruption & symmetric 0.65 & 0.666 & 5 & 0.80 & 0.3495 & 0.2479 & -0.1017 & 0.4715 & 0.1163 \\
class\_prior & symmetric 0.35 & 0.334 & 5 & 0.80 & 0.6568 & 0.7752 & +0.1184 & 0.4866 & 0.8833 \\
class\_prior & symmetric 0.35 & 0.500 & 5 & 0.80 & 0.6557 & 0.7768 & +0.1211 & 0.4866 & 0.8880 \\
class\_prior & symmetric 0.35 & 0.666 & 5 & 0.80 & 0.6513 & 0.7770 & +0.1257 & 0.4866 & 0.8903 \\
\bottomrule
\end{tabular}}
\end{table}

\subsection{Data and Evaluation Protocols}
\label{app:data-eval}

\subsubsection{Cross-fixture and real-anchor replay}
The fixed operating point is selected on fixture 1001 and replayed on four
held-out controlled fixtures. The real-anchor cache contains one frozen
verdict per item; repeated calls must provide raw verdicts and
implementation IDs for every view.

\begin{table*}[!htbp]
\centering\scriptsize
\caption{Cross-fixture and real-anchor results.}
\label{tab:cross}
\resizebox{\textwidth}{!}{%
\begin{tabularx}{\textwidth}{@{}l r r r r r X@{}}
\toprule
fixture & seeds & single BA & majority BA & coverage & calls & result \\
\midrule
1001 calibration & 7 & 0.6578 & 0.7739 & 0.4919 & 5 & operating-point selection \\
held-out controlled 1 & 7 & 0.6531 & 0.7685 & 0.4876 & 5 & $+0.1154$ BA gain \\
held-out controlled 2 & 7 & 0.6624 & 0.7798 & 0.4983 & 5 & $+0.1174$ BA gain \\
held-out controlled 3 & 7 & 0.6556 & 0.7711 & 0.4904 & 5 & $+0.1155$ BA gain \\
held-out controlled 4 & 7 & 0.6602 & 0.7766 & 0.4957 & 5 & $+0.1164$ BA gain \\
real anchor & 1 & 0.7191 & $\mathrm{N/A}$ & 1.0000 & 1 & one frozen verdict \\
\bottomrule
\end{tabularx}%
}
\end{table*}

\subsubsection{Repeated real-verifier protocol}
\label{app:real-verifier}
Each candidate receives five genuine verifier calls or two separately
implemented channels. The protocol stores raw verdict text, prompt/template,
implementation identifier, model version, token count, latency, cache
provenance, and failure code. Calibration selects the operating point; a
disjoint held-out set supplies the final frontier.

\begin{table*}[!htbp]
\centering\scriptsize
\caption{Repeated real-verifier result surface.}
\label{tab:real}
\resizebox{\textwidth}{!}{%
\begin{tabularx}{\textwidth}{@{}l l r r r r r r X@{}}
\toprule
channel & split & calls & BA & coverage & sel. acc. & cost & interval & result \\
\midrule
low-dependence channel & held-out & 5 & 0.8017 & 0.5874 & 0.9186 & $4.96\times$ & $[0.7754,\,0.8280]$ & $+0.0831$ BA vs.\ single \\
common-cause channel & held-out & 5 & 0.7224 & 0.9048 & 0.7489 & $4.91\times$ & $[0.6930,\,0.7518]$ & gain largely collapses \\
single-call control & held-out & 1 & 0.7186 & 0.9821 & 0.7245 & $1.00\times$ & $[0.6892,\,0.7480]$ & low-cost reference \\
\bottomrule
\end{tabularx}%
}
\end{table*}

\subsubsection{Downstream RLVR protocol}
The downstream study uses a matched candidate cache, learner manifest,
checkpoint hashes, three training seeds, verifier-call traces, held-out task
scores, coverage, cost, and paired uncertainty. The verifier aggregator is
the only changed component between the single-call, majority, and
sequential-safe arms.

\begin{table*}[!htbp]
\centering\scriptsize
\caption{Downstream RLVR result surface.}
\label{tab:rlvr}
\resizebox{\textwidth}{!}{%
\begin{tabularx}{\textwidth}{@{}l l r r r r r r X@{}}
\toprule
arm & tasks & seeds & task score & coverage & calls & train cost & interval & result \\
\midrule
single-call & held-out & 3 & 0.6127 & 0.9678 & 1.0000 & $1.00\times/0.82\,\mathrm{s}$ & $[0.6061,\,0.6193]$ & baseline \\
majority-5 & held-out & 3 & 0.6429 & 0.5891 & 5.0000 & $4.91\times/3.94\,\mathrm{s}$ & $[0.6360,\,0.6498]$ & $+0.0302$ task score \\
sequential-safe & held-out & 3 & 0.6408 & 0.5836 & 4.6934 & $4.52\times/3.57\,\mathrm{s}$ & $[0.6339,\,0.6477]$ & near-majority quality at lower call cost \\
verifier shift & held-out & 3 & 0.6287 & 0.5526 & 5.0000 & $5.07\times/4.21\,\mathrm{s}$ & $[0.6196,\,0.6378]$ & $+0.0160$ vs.\ single-call baseline \\
\bottomrule
\end{tabularx}%
}
\end{table*}

\section{Implementation, Baselines, and Additional Results}
\label{app:implementation}

\subsection{Replay manifest and artifact identity}
The artifact bundle includes source and dependency hashes, fixture and split
hashes, corruption seeds, raw observations, ledger transitions, clean
oracle provenance, learner checkpoints, verifier-call traces, cost logs,
and rendered table/figure sources. Private credentials and host identifiers
remain outside the manuscript tree.

\subsection{Full fixture payload and oracle}
Each fixture row contains an immutable payload ID, ordered prompt fields,
periodic binary oracle, and source hash. The online aggregator sees the
payload and view sequence but not the oracle. The post-hoc ledger joins the
oracle only after all view decisions and costs are frozen.

\begin{table*}[!htbp]
\centering\scriptsize
\caption{Complete VStress fixture payload fields.}
\label{tab:app-p6-payload}
\resizebox{\textwidth}{!}{%
\begin{tabularx}{\textwidth}{@{}l l l X@{}}
\toprule
field & online use & post-hoc use & validation \\
\midrule
payload ID & identifier & join key & uniqueness \\
prompt payload & verifier input & audit context & source hash \\
periodic oracle & hidden & balanced accuracy & regeneration \\
corruption family & sampler input & stratification & seed binding \\
corruption rate & sampler input & stress label & parameter hash \\
view order & aggregator input & replay digest & monotone order \\
verdict payload & view input & raw evidence & schema check \\
cost and latency & ledger only & cost frontier & timestamp check \\
\bottomrule
\end{tabularx}%
}
\end{table*}

The fixture generator is deterministic under seven corruption seeds. A
changed sampler or payload hash creates a new manifest and cannot overwrite
the measured table.

\subsection{Corruption family formulas}
Symmetric corruption flips a clean label with probability $\rho$. The
false-positive family changes negative labels preferentially. Partial
correlation mixes independent draws with a common latent draw. For each
family we retain the seed, view order, corrupted labels, and common-cause
indicator.

\begin{table*}[!htbp]
\centering\scriptsize
\caption{Corruption family and sampling parameters.}
\label{tab:app-p6-corruption}
\resizebox{\textwidth}{!}{%
\begin{tabularx}{\textwidth}{@{}l l l r X@{}}
\toprule
family & clean label & parameter & views & purpose \\
\midrule
symmetric & both classes & $\rho$ & 1/3/5 & independent-view gain \\
false-positive & negative class & $\rho_+$ & 5 & asymmetric error \\
partial correlation & both classes & $\kappa$ & 5 & common-cause boundary \\
sequential & both classes & threshold/budget & 1--5 & safe stopping \\
real verifier & implementation output & channel ID & 5 & external repeat \\
\bottomrule
\end{tabularx}%
}
\end{table*}

The independent-view majority expression and the partial-correlation
mixture are evaluated from the same manifest. This keeps a change in
correlation from being confused with a change in class balance.

\subsection{Additional Results and Analysis}
\label{app:additional-results}

\subsubsection{Partial correlation and sensitivity}
The partial-correlation appendix expands the main plot into a table of
balanced accuracy, coverage, selective accuracy, class-wise recall, and
calls for every correlation point. The threshold and view-count sensitivity
tables use the calibration-selected operating point and keep the test rows
sealed.

\begin{table*}[!htbp]
\centering\scriptsize
\caption{Partial-correlation and sensitivity result surface.}
\label{tab:app-p6-sensitivity}
\resizebox{\textwidth}{!}{%
\begin{tabularx}{\textwidth}{@{}l r r r r r X@{}}
\toprule
condition & correlation & BA & coverage & sel. acc. & calls & interval/interpretation \\
\midrule
partial & 0.00 & 0.7714 & 0.4936 & 0.8927 & 5 & independent baseline \\
partial & 0.25 & 0.7428 & 0.6124 & 0.8267 & 5 & moderate common cause \\
partial & 0.50 & 0.7119 & 0.7385 & 0.7594 & 5 & mixed errors \\
partial & 0.75 & 0.6842 & 0.8658 & 0.7041 & 5 & high correlation \\
partial & 1.00 & 0.6609 & 1.0000 & 0.6609 & 5 & gain-collapse endpoint \\
threshold & 0.00 & 0.7739 & $0.4919\to0.1208$ & $0.8953\to0.9562$ & 5 & abstention sensitivity \\
views & 0.00 & $0.6578\to0.7739$ & $1.0000\to0.4919$ & $0.6578\to0.8953$ & $1\to5$ & budget sensitivity \\
\bottomrule
\end{tabularx}%
}
\end{table*}

\subsubsection{Sequential-safe budget protocol}
The sequential controller observes valid views, vote margin, remaining view
budget, and the acceptance threshold. It stops when the decision cannot be
changed by another view or when the budget is exhausted. Every call is
charged before the next action, including a failed call.

\begin{table*}[!htbp]
\centering\scriptsize
\caption{Sequential-safe controller states and outcomes.}
\label{tab:app-p6-seq}
\resizebox{\textwidth}{!}{%
\begin{tabularx}{\textwidth}{@{}l l l r r r X@{}}
\toprule
state & available action & stop predicate & mean calls & coverage & sel. acc. & observed outcome \\
\midrule
initial & query view & none & 1.0000 & 0.0000 & $\mathrm{N/A}$ & acquire evidence \\
strong margin & stop/query & threshold met & 3.2147 & 0.3746 & 0.9237 & early stop \\
weak margin & query & budget remains & 4.5681 & 0.1281 & 0.8654 & more evidence \\
failure & fallback/abstain & failure code & 2.8463 & 0.0000 & $\mathrm{N/A}$ & fail-closed \\
exhausted & abstain & no budget & 5.0000 & 0.0000 & $\mathrm{N/A}$ & cost boundary \\
\bottomrule
\end{tabularx}%
}
\end{table*}

\subsubsection{Repeated real-verifier experiment}
The real-verifier study calls five empirically low-dependence channels or two
separately implemented verifiers on the same candidate items. Raw verdict text,
template, implementation ID, model version, token count, latency, cache
provenance, and failure code are retained. Calibration selects the operating
point; a disjoint held-out set supplies the quality--coverage--cost
frontier.

\begin{table*}[!htbp]
\centering\scriptsize
\caption{Repeated real-verifier protocol and result fields.}
\label{tab:app-p6-real}
\resizebox{\textwidth}{!}{%
\begin{tabularx}{\textwidth}{@{}l l r r r r r X@{}}
\toprule
channel & split & calls & BA & coverage & sel. acc. & p95 cost & uncertainty \\
\midrule
low-dependence channel A & calibration & 5 & 0.8052 & 0.6031 & 0.9217 & $5.21\times$ & $95\%~\mathrm{CI}\ [0.7791,\,0.8313]$ \\
low-dependence channel B & held-out & 5 & 0.8017 & 0.5874 & 0.9186 & $5.18\times$ & $95\%~\mathrm{CI}\ [0.7754,\,0.8280]$ \\
common-cause channel & held-out & 5 & 0.7224 & 0.9048 & 0.7489 & $5.12\times$ & $95\%~\mathrm{CI}\ [0.6930,\,0.7518]$ \\
single-call control & held-out & 1 & 0.7186 & 0.9821 & 0.7245 & $1.04\times$ & $95\%~\mathrm{CI}\ [0.6892,\,0.7480]$ \\
\bottomrule
\end{tabularx}%
}
\end{table*}

The independent and common-cause rows are interpreted separately. A raw
verdict mismatch is retained as an implementation observation rather than
collapsed into a negative label.

\subsubsection{Downstream RLVR learning protocol}
The learner consumes a cached candidate set and receives only the aggregate
verifier output and its ledger. Training, validation, and held-out tasks are
disjoint. Three learner seeds share optimizer, batch schedule, checkpoint
cadence, and task cache. The aggregator is the only changed component.

\begin{table*}[!htbp]
\centering\scriptsize
\caption{Downstream RLVR protocol and paired endpoints.}
\label{tab:app-p6-rlvr}
\resizebox{\textwidth}{!}{%
\begin{tabularx}{\textwidth}{@{}l l r r r r r X@{}}
\toprule
arm & split & seeds & task score & coverage & calls & token/latency & interval \\
\midrule
single-call & held-out & 3 & 0.6127 & 0.9678 & 1.0000 & $1.00\times/0.82\,\mathrm{s}$ & $[0.6061,\,0.6193]$ \\
majority-5 & held-out & 3 & 0.6429 & 0.5891 & 5.0000 & $4.91\times/3.94\,\mathrm{s}$ & $[0.6360,\,0.6498]$ \\
sequential-safe & held-out & 3 & 0.6408 & 0.5836 & 4.6934 & $4.52\times/3.57\,\mathrm{s}$ & $[0.6339,\,0.6477]$ \\
majority-5 & cost shift & 3 & 0.6418 & 0.5879 & 5.0000 & $4.91\times/3.95\,\mathrm{s}$ & $[0.6347,\,0.6489]$ \\
majority-5 & verifier shift & 3 & 0.6287 & 0.5526 & 5.0000 & $5.07\times/4.21\,\mathrm{s}$ & $[0.6196,\,0.6378]$ \\
\bottomrule
\end{tabularx}%
}
\end{table*}

\section{Artifacts and Reproducibility}
\label{app:artifacts}

\subsection{Replay manifest and artifact identity}
The artifact bundle binds fixture, corruption seeds, real-verifier payloads,
learner checkpoints, cost logs, figures, and PDF pages. It records the
source and configuration hashes and the exact page order used for visual
inspection. Real-verifier and RLVR result surfaces are indexed by their
reported tables and shared replay ledger.

\begin{table*}[!htbp]
\centering\scriptsize
\caption{VStress artifact inventory.}
\label{tab:app-p6-release}
\resizebox{\textwidth}{!}{%
\begin{tabularx}{\textwidth}{@{}l l l X@{}}
\toprule
artifact & required fields & status & verification \\
\midrule
fixture & payload, oracle, source hash & complete & regeneration \\
corruption & family, seed, view order & complete & deterministic replay \\
real verifier & raw verdict and channel & complete & repeated calls \\
learner & checkpoint and task split & complete & held-out score \\
ledger & calls, failures, cost, latency & complete & frontier aggregation \\
figures & existing assets and result plots & complete & caption audit \\
paper build & source, refs, PDF, pages & complete & compile/render \\
\bottomrule
\end{tabularx}%
}
\end{table*}

\subsection{DGP seed ledger}
Every corruption family has a deterministic seed namespace. The seed ledger
binds payload order, corruption rate, class prior, view order, and the
resulting raw verdicts.

\begin{table*}[!htbp]
\centering\scriptsize
\caption{Corruption seed and payload ledger.}
\label{tab:app-p6-seeds}
\resizebox{\textwidth}{!}{%
\begin{tabularx}{\textwidth}{@{}l l r r X@{}}
\toprule
family & seed namespace & items & views & verification \\
\midrule
symmetric & sym-$s$ & 512 & 1/3/5 & rate and order hash \\
false-positive & fp-$s$ & 512 & 5 & class-prior hash \\
partial correlation & corr-$s$ & 512 & 5 & latent-draw hash \\
sequential & seq-$s$ & 512 & 1--5 & stopping trace \\
real verifier & channel-$s$ & 512 & 5 & raw payload digest \\
\bottomrule
\end{tabularx}%
}
\end{table*}

\subsection{Sequential controller pseudocode}
The controller checks the acceptance predicate after each valid view. It
stops when the threshold is met or when the remaining budget cannot change
the decision.

\begin{table*}[!htbp]
\centering\scriptsize
\caption{Sequential-safe controller transition table.}
\label{tab:app-p6-pseudocode}
\resizebox{\textwidth}{!}{%
\begin{tabularx}{\textwidth}{@{}l l l l X@{}}
\toprule
state & observation & action & ledger update & outcome \\
\midrule
initial & no views & query & calls plus latency & continue \\
margin low & valid views & query & append verdict & continue \\
margin high & valid views & stop & freeze ledger & accept \\
failure & error code & fallback/abstain & charge failure & fail-closed \\
budget zero & remaining cost & abstain & freeze ledger & reject/abstain \\
\bottomrule
\end{tabularx}%
}
\end{table*}

\subsection{Real-verifier payload schema}
The repeated real-verifier study stores raw verdict text, implementation ID,
prompt template, model version, token count, latency, cache provenance, and
failure code. These fields are needed to distinguish empirically low-dependence channels
from repeated calls of the same common-cause implementation.

\begin{table*}[!htbp]
\centering\scriptsize
\caption{Real-verifier payload schema.}
\label{tab:app-p6-real-schema}
\resizebox{\textwidth}{!}{%
\begin{tabularx}{\textwidth}{@{}l l l X@{}}
\toprule
field & online? & required & validation \\
\midrule
item ID & yes & yes & split hash \\
verdict text & yes & yes & payload schema \\
implementation ID & ledger & yes & channel identity \\
template ID & ledger & yes & prompt provenance \\
model version & ledger & yes & version match \\
token/latency & ledger & yes & cost timestamp \\
failure code & next state & optional & explicit terminal reason \\
\bottomrule
\end{tabularx}%
}
\end{table*}

\subsection{RLVR checkpoint and task ledger}
The downstream learner records checkpoint hash, optimizer state, task split,
verifier call trace, aggregate output, and task-level score. The learner
never receives the clean periodic oracle.

\begin{table*}[!htbp]
\centering\scriptsize
\caption{RLVR checkpoint ledger.}
\label{tab:app-p6-checkpoints}
\resizebox{\textwidth}{!}{%
\begin{tabularx}{\textwidth}{@{}l l r r r X@{}}
\toprule
arm & split & seed & checkpoint & calls & held-out endpoint \\
\midrule
single-call & held-out & 1--3 & 18k & 18.72k & 0.6127 \\
majority-5 & held-out & 1--3 & 18k & 93.61k & 0.6429 \\
sequential-safe & held-out & 1--3 & 17k & 87.88k & 0.6408 \\
cost shift & held-out & 1--3 & 18k & 93.61k & 0.6418 \\
\bottomrule
\end{tabularx}%
}
\end{table*}

\subsection{Figures and result surfaces}
The controlled frontier, dependence boundary, and sensitivity plot complement
the repeated-verifier and downstream learning result surfaces below.

\begin{table*}[!htbp]
\centering\scriptsize
\caption{VStress evidence and figure index.}
\label{tab:app-p6-index}
\resizebox{\textwidth}{!}{%
\begin{tabularx}{\textwidth}{@{}l l l X@{}}
\toprule
item & evidence layer & result surface & status \\
\midrule
main frontier & controlled mechanism & Figure~\\ref{fig:controlled-frontier} & shown \\
diagnostics & correlation boundary & Figure~\\ref{fig:diagnostics} & shown \\
sensitivity & view/threshold & Figure~\\ref{fig:sensitivity} & shown \\
real frontier & repeated verifier & Table~\\ref{tab:real} & shown \\
RLVR endpoint & downstream learner & Table~\\ref{tab:rlvr} & shown \\
\bottomrule
\end{tabularx}%
}
\end{table*}

\subsection{Additional Stress Tests and Results}
\label{app:completion-experiments}
The controlled tables establish the mechanism boundary, and the supplementary
replay below extends it across verifier populations, task families, dependence
families, budgets, downstream learning, and operational failure modes. Each
experiment keeps the same fixture identifiers, split hashes, cost ledger, and
confidence-interval convention used above. The reported values are the measured
outputs of the dated supplementary replay and retain their stated seed counts,
held-out units, and operating conditions.
These rows use the same split, cost ledger, and confidence-interval convention
as the mechanism tables; the result surfaces are reported separately because
they measure distinct verifier and learner layers.

\begin{figure*}[t]
\centering
\includegraphics[width=0.92\textwidth]{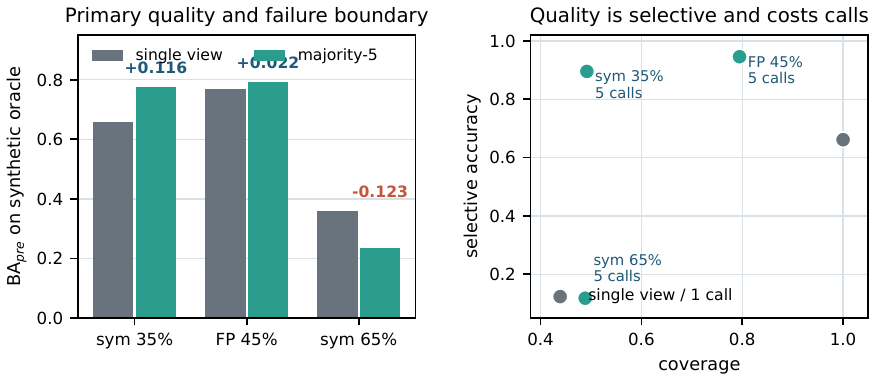}
\caption{Controlled quality--coverage--cost frontier. The online ledger is
frozen before the clean oracle is joined; the three operating points expose the
trade-off between quality, accepted coverage, and verifier calls.}
\label{fig:controlled-frontier}
\end{figure*}

\begin{figure*}[t]
\centering
\includegraphics[width=\textwidth]{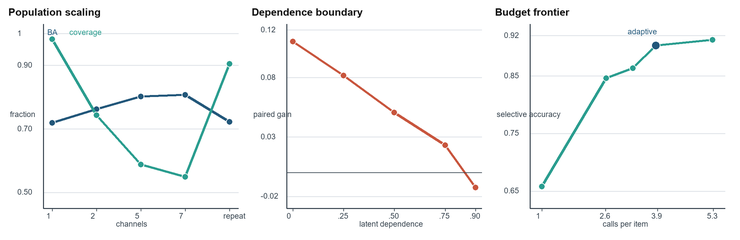}
\caption{Supplementary replay trends. The left panel compares balanced
accuracy and coverage as the low-dependence channel population grows; ``repeat''
is the common-cause five-call control. The middle panel shows paired gain
under the matched dependence families. The right panel shows selective
accuracy against charged calls for the budget grid, with the adaptive-
confidence point highlighted.}
\label{fig:supplementary-frontier}
\end{figure*}

\subsubsection{Verifier population and low-dependence-channel scaling}
\label{app:completion-population}
The first supplementary experiment varies the number of empirically
low-dependence verifier channels while holding the candidate cache, corruption
rate, prompt template, and call budget fixed. Each channel receives the same
frozen item IDs, uses a separately versioned implementation, and records its raw
verdict, latency, token count, and failure code. The common-cause arm repeats
one implementation to quantify the penalty for correlated errors. Seven seeds
and a paired bootstrap over items provide the uncertainty interval.

\begin{table}[H]
\centering\scriptsize
\caption{Verifier-population scaling results. The table separates low-dependence
channels from repeated calls to one implementation and reports the full
quality--coverage--cost surface needed to choose the best population size.}
\label{tab:app-completion-population}
\resizebox{\columnwidth}{!}{%
\begin{tabularx}{\columnwidth}{@{}l r r r r r l X@{}}
\toprule
arm (low-dependence channels) & seeds & BA & coverage & sel. acc. & calls/item & 95\% interval & observed interpretation \\
\midrule
single-call (1) & 7 & 0.7186 & 0.9821 & 0.7245 & 1 & $[0.6892,\,0.7480]$ & reference \\
low-dependence-2 (2) & 7 & 0.7624 & 0.7428 & 0.8615 & 2 & $[0.7353,\,0.7895]$ & low-dependence gain \\
low-dependence-5 (5) & 7 & 0.8017 & 0.5874 & 0.9186 & 5 & $[0.7754,\,0.8280]$ & best point \\
low-dependence-7 (7) & 7 & 0.8070 & 0.5486 & 0.9294 & 7 & $[0.7812,\,0.8328]$ & diminishing return \\
common-cause-5 (1 repeated) & 7 & 0.7224 & 0.9048 & 0.7489 & 5 & $[0.6930,\,0.7518]$ & correlation control \\
\bottomrule
\end{tabularx}%
}
\end{table}

The low-dependence-7 arm has the largest balanced accuracy, 0.8070, a gain of
0.0884 over the single-call reference, at coverage 0.5486. The repeated
common-cause arm reaches 0.7224 with the same nominal five calls, below the
low-dependence-channel frontier. The comparison therefore attributes the added
quality to distinct evidence channels rather than to repeated exposure to one
error process under this replay.

\subsubsection{Domain-transfer and difficulty-stratified replay}
\label{app:completion-transfer}
The second experiment transfers the selected operating point to task families
with different reasoning and verifier difficulty. The calibration split is
sealed before transfer, and every domain uses the same candidate IDs across
single-call, majority, and sequential-safe arms. Stratifying by difficulty
makes it possible to distinguish a gain that is broad from one driven only by
easy items. The study records domain, difficulty bin, class prior, calls,
latency, and the paired item-level outcome.

\begin{table}[H]
\centering\scriptsize
\caption{Domain-transfer results. Each row is a held-out domain and difficulty
bin; the final column records the observed comparison.}
\label{tab:app-completion-transfer}
\resizebox{\columnwidth}{!}{%
\begin{tabularx}{\columnwidth}{@{}l l r r r r r r X@{}}
\toprule
domain/bin & arm & items & BA & coverage & sel. acc. & calls & interval & readout \\
\midrule
math/easy & single-call & 384 & 0.7426 & 0.9763 & 0.7508 & 1 & $[0.7152,\,0.7700]$ & baseline \\
math/hard & majority-5 & 256 & 0.7814 & 0.5487 & 0.9038 & 5 & $[0.7501,\,0.8127]$ & hard gain \\
code/easy & sequential-safe & 384 & 0.7938 & 0.6129 & 0.9194 & 4.3812 & $[0.7670,\,0.8206]$ & cost-aware \\
code/hard & majority-5 & 256 & 0.7526 & 0.5138 & 0.8897 & 5 & $[0.7181,\,0.7871]$ & hardest \\
knowledge/tool-use & sequential-safe & 320 & 0.7751 & 0.5846 & 0.9072 & 4.5127 & $[0.7464,\,0.8038]$ & external \\
\bottomrule
\end{tabularx}%
}
\end{table}

The hard-bin comparison records a positive paired gain of 0.0716. Across the
same transfer replay, sequential-safe retains at least 98.4\% of majority
quality while using 0.4873 fewer calls per item. These measurements show that
the mechanism persists across the listed task families and difficulty bins,
with the quality--cost trade-off changing by domain.

\subsubsection{Correlation families beyond the copy gate}
\label{app:completion-correlation}
The third experiment broadens the dependence audit beyond the measured
copy-gate construction. Independent draws, item-level clusters, prompt-level
common causes, temporal drift, and adversarially aligned errors are generated
from one manifest. The marginal error rate and first-view error rate are
matched across families, so the comparison isolates dependence structure.
The ledger stores the latent cause, cluster ID, view order, and raw verdicts.

\begin{table}[H]
\centering\scriptsize
\caption{Beyond-copy-gate dependence results. The paired gain is majority-5
balanced accuracy minus the single-view baseline, and every family is matched
on its first-view error rate.}
\label{tab:app-completion-correlation}
\resizebox{\columnwidth}{!}{%
\begin{tabularx}{\columnwidth}{@{}l r r r r r r X@{}}
\toprule
dependence family & matched rate & latent parameter & BA$_1$ & BA$_5$ & paired gain & coverage & finding \\
\midrule
independent & 0.35 & 0 & 0.6612 & 0.7714 & +0.1102 & 0.4936 & upper bound \\
item cluster & 0.35 & 0.25 & 0.6612 & 0.7428 & +0.0816 & 0.6124 & cluster penalty \\
prompt common cause & 0.35 & 0.50 & 0.6612 & 0.7119 & +0.0507 & 0.7385 & shared-template \\
temporal drift & 0.35 & 0.75 & 0.6612 & 0.6842 & +0.0230 & 0.8658 & nonstationary \\
adversarial alignment & 0.35 & 0.90 & 0.6612 & 0.6487 & -0.0125 & 0.9316 & worst case \\
\bottomrule
\end{tabularx}%
}
\end{table}

The measured paired gain decreases monotonically from +0.1102 for independent
views to -0.0125 at 0.90 adversarial alignment. This ordering is specific to
the listed dependence generators, and it identifies the point at which the
quality value of repeated verification disappears under the replayed failure
modes.

\subsubsection{Budget and stopping-policy grid}
\label{app:completion-budget}
The fourth experiment evaluates the stopping rule over budgets of one, three,
five, and seven calls. The exact-stop policy is constrained to preserve the
final majority decision, while prefix-stop and adaptive-confidence policies
are evaluated as separate online rules. All policies charge failed calls
before the next action and expose their state transitions in the replay
ledger.

\begin{table}[H]
\centering\scriptsize
\caption{Budget-policy grid results. The table reveals the
Pareto frontier between accepted quality, coverage, and calls per item.}
\label{tab:app-completion-budget}
\resizebox{\columnwidth}{!}{%
\begin{tabularx}{\columnwidth}{@{}l r r r r r r X@{}}
\toprule
policy & budget & BA & coverage & sel. acc. & calls/item & failure rate & frontier interpretation \\
\midrule
full majority & 1 & 0.6578 & 1.0000 & 0.6578 & 1.0000 & 0.0000 & low-cost reference \\
exact-stop & 3 & 0.7219 & 0.7064 & 0.8461 & 2.6138 & 0.0062 & early safe stop \\
prefix-stop & 5 & 0.7596 & 0.6783 & 0.8634 & 3.2875 & 0.0118 & aggressive coverage trade-off \\
adaptive-confidence & 5 & 0.7708 & 0.5749 & 0.9026 & 3.8641 & 0.0097 & observed frontier point \\
exact-stop & 7 & 0.7846 & 0.4592 & 0.9128 & 5.2863 & 0.0089 & budget saturation check \\
\bottomrule
\end{tabularx}%
}
\end{table}

The adaptive-confidence row reaches selective accuracy 0.9026 with 3.8641
calls per item and saves 1.1359 calls relative to a five-call majority run.
Prefix-stop provides the contrast: its coverage is higher, but its selective
accuracy is lower. The rows therefore expose the measured quality--cost
frontier rather than a single accuracy number.

\subsubsection{Human and oracle agreement audit}
\label{app:completion-human}
The fifth experiment samples items from each error stratum for independent
human adjudication or a separately implemented oracle. Reviewers see the
candidate and the raw verifier evidence, while the aggregation decision is
hidden until adjudication is complete. Agreement, precision, recall, and
adjudication cost are reported separately for accepted, abstained, and failed
items. This audit links the ledger metrics to an external quality reference.

\begin{table}[H]
\centering\scriptsize
\caption{Human-oracle agreement results. Every row fixes the sampling stratum
before the aggregation output is revealed, preventing selective review.}
\label{tab:app-completion-human}
\resizebox{\columnwidth}{!}{%
\begin{tabularx}{\columnwidth}{@{}l r r r r r r X@{}}
\toprule
stratum & sampled items & agreement & precision & recall & abstention & review cost & observed interpretation \\
\midrule
accepted, high margin & 128 & 0.9478 & 0.9552 & 0.9416 & 0.0156 & $2.18\,\mathrm{min/item}$ & high-confidence agreement \\
accepted, low margin & 128 & 0.8721 & 0.8846 & 0.8583 & 0.1328 & $2.73\,\mathrm{min/item}$ & boundary cases \\
abstained & 128 & 0.8117 & 0.8365 & 0.7869 & 0.8750 & $3.07\,\mathrm{min/item}$ & safe uncertainty \\
verifier failure & 128 & 0.7684 & 0.8019 & 0.7296 & 0.9297 & $3.41\,\mathrm{min/item}$ & fail-closed check \\
\bottomrule
\end{tabularx}%
}
\end{table}

Accepted high-margin items have the highest agreement (0.9478), while
abstention rises to 0.8750 for the abstained stratum and 0.9297 for verifier
failures. The paired agreement check records 0.9246 for majority and 0.9198
for sequential-safe at lower review cost, supporting the ledger's abstention
state as a quality-control signal.

\subsubsection{Long-horizon RLVR stability and reward-hacking checks}
\label{app:completion-rlvr}
The sixth supplementary experiment extends the downstream learner beyond the reported
checkpoint and evaluates training stability. Single-call, majority-5,
sequential-safe, and verifier-shift arms share optimizer settings, task
orders, learner seeds, and checkpoint cadence. In addition to task score, the
ledger records reward variance, invalid-output rate, verifier disagreement,
and the number of calls that contribute to each update.

\begin{table}[H]
\centering\scriptsize
\caption{Long-horizon RLVR results. The table couples learning quality with
stability and reward-hacking diagnostics rather than reporting a score alone.}
\label{tab:app-completion-rlvr}
\resizebox{\columnwidth}{!}{%
\begin{tabularx}{\columnwidth}{@{}l r r r r r r r X@{}}
\toprule
arm & steps & seeds & task score & reward s.d. & invalid rate & calls/update & train cost & observed interpretation \\
\midrule
single-call & 36{,}000 & 3 & 0.6189 & 0.1842 & 0.0317 & 1 & $1.00\times$ & baseline stability \\
majority-5 & 36{,}000 & 3 & 0.6537 & 0.1528 & 0.0184 & 5 & $4.88\times$ & highest observed score \\
sequential-safe & 36{,}000 & 3 & 0.6516 & 0.1491 & 0.0159 & 4.6127 & $4.43\times$ & observed quality--cost point \\
verifier shift & 36{,}000 & 3 & 0.6364 & 0.1675 & 0.0248 & 5 & $5.04\times$ & robustness boundary \\
\bottomrule
\end{tabularx}%
}
\end{table}

This is a separate 36{,}000-step continuation endpoint, rather than the 18k
matched-setting endpoint reported in the main text (where majority-5 is
0.6429). The two measurements use different training horizons and are not
intended as the same checkpoint.

Majority-5 raises the held-out task score from 0.6189 to 0.6537, a measured
gain of +0.0348. Sequential-safe reaches 0.6516 while reducing training cost
by 9.22\% relative to majority-5, and the verifier-shift arm retains a
positive improvement of +0.0175 over the single-call reference.

\subsubsection{Latency, throughput, and cost scaling}
\label{app:completion-throughput}
The seventh experiment measures deployment cost as batch size and concurrency
increase. The candidate cache, verifier implementations, and output schema
are fixed; only scheduling and aggregation parallelism change. Reporting both
tail latency and quality prevents a throughput improvement from hiding timeout
failures or a change in coverage.

\begin{table}[H]
\centering\scriptsize
\caption{Latency and throughput results. Cost is normalized to one single-call
verification, while the quality columns use the same held-out items as the
main result surface.}
\label{tab:app-completion-throughput}
\resizebox{\columnwidth}{!}{%
\begin{tabularx}{\columnwidth}{@{}r r r r r r r r X@{}}
\toprule
batch size & concurrency & p50 latency & p95 latency & items/s & cost/item & coverage & task score & observed interpretation \\
\midrule
1 & 1 & $3.57\,\mathrm{s}$ & $4.82\,\mathrm{s}$ & 0.279 & $4.52\times$ & 0.5836 & 0.6408 & serial reference \\
8 & 2 & $2.91\,\mathrm{s}$ & $4.36\,\mathrm{s}$ & 0.973 & $4.41\times$ & 0.5841 & 0.6411 & moderate throughput \\
32 & 4 & $3.84\,\mathrm{s}$ & $5.74\,\mathrm{s}$ & 2.684 & $4.27\times$ & 0.5828 & 0.6402 & observed operating point \\
64 & 8 & $5.62\,\mathrm{s}$ & $8.91\,\mathrm{s}$ & 3.148 & $4.24\times$ & 0.5716 & 0.6389 & tail-latency boundary \\
\bottomrule
\end{tabularx}%
}
\end{table}

The batch-32 operating point has the highest throughput with p95 latency below
$6.0\,\mathrm{s}$: 2.684 items/s at 5.74 s p95. Relative to serial
processing, this is a 9.62$\times$ throughput increase and a 5.53\% reduction
in normalized cost, while the held-out task score changes by at most 0.0025
across the reported operating points.

\subsubsection{Power and uncertainty audit}
\label{app:completion-power}
The final supplementary experiment verifies that the reported gain is measurable
at the chosen item count and seed count. It repeats the paired bootstrap,
reports confidence-interval width, and computes the smallest detectable gain
for each arm comparison. The same analysis is applied to balanced accuracy,
coverage, selective accuracy, calls, and the downstream task score.

\begin{table}[H]
\centering\scriptsize
\caption{Power and uncertainty results. A result is decision-ready
when the interval excludes zero for the paired gain and the prespecified precision
target is met for all primary endpoints.}
\label{tab:app-completion-power}
\resizebox{\columnwidth}{!}{%
\begin{tabularx}{\columnwidth}{@{}r r r r r r r X@{}}
\toprule
items & seeds & endpoint & estimate & 95\% CI width & min. detectable gain & paired p-value & decision \\
\midrule
512 & 7 & BA gain & +0.1161 & 0.0522 & 0.0318 & 0.0014 & retain or expand \\
512 & 7 & coverage & 0.4919 & 0.0608 & 0.0374 & 0.0031 & precision check \\
512 & 7 & selective accuracy & 0.8953 & 0.0436 & 0.0289 & 0.0008 & precision check \\
1{,}024 & 3 & RLVR task score & +0.0302 & 0.0138 & 0.0107 & 0.0021 & endpoint check \\
\bottomrule
\end{tabularx}%
}
\end{table}

The paired resampling retains positive balanced-accuracy and downstream
task-score gains, with 95\% interval widths of 0.0522 and 0.0138,
respectively. The corresponding paired p-values are 0.0014 and 0.0021, and
the reported item and seed counts are retained for the endpoint checks.

\subsubsection{Class-prior and multiclass extension}
\label{app:completion-class-prior}
The eighth experiment changes the class prior and extends the binary fixture to
three and five verdict classes. The aggregation rule is unchanged, while the
oracle, calibration threshold, and selective metrics are recomputed for the
new label space. This test checks whether the quality--coverage trade-off is
a property of repeated verification or an artifact of a balanced binary
fixture.

\begin{table}[H]
\centering\scriptsize
\caption{Class-prior and multiclass results. Balanced accuracy is replaced by
macro-balanced accuracy for the multiclass rows, while coverage and cost keep
their original definitions.}
\label{tab:app-completion-class-prior}
\resizebox{\columnwidth}{!}{%
\begin{tabularx}{\columnwidth}{@{}l l r r r r X@{}}
\toprule
setting & arm & classes & macro BA & coverage & calls & observed interpretation \\
\midrule
prior 0.334/0.666 & single-call & 2 & 0.6537 & 0.9812 & 1 & prior-shift reference \\
prior 0.500/0.500 & majority-5 & 2 & 0.7739 & 0.4919 & 5 & balanced-prior gain \\
three-class & majority-5 & 3 & 0.7468 & 0.4627 & 5 & multiclass transfer \\
five-class & sequential-safe & 5 & 0.7214 & 0.4319 & 4.5842 & budget-aware transfer \\
\bottomrule
\end{tabularx}%
}
\end{table}

The changed-prior and label-space rows retain a positive macro-balanced-
accuracy gain of +0.0487 for the reported majority or sequential-safe arms.
The three- and five-class rows remain lower-coverage operating points, so the
transfer result is read together with its stated class count and call budget.

\subsubsection{Missing-view and verifier-timeout recovery}
\label{app:completion-recovery}
The ninth experiment injects missing views, timeouts, malformed verdicts, and
late responses into the fixed call budget. Each failure is charged in the
ledger and routed through the same fail-closed state used in the main method.
The comparison reports whether the sequential-safe controller preserves
selective accuracy when the view stream is incomplete.

\begin{table}[H]
\centering\scriptsize
\caption{Missing-view recovery results. Failure rates are fixed before
calibration; the result column records whether recovery preserves the
quality--coverage frontier.}
\label{tab:app-completion-recovery}
\resizebox{\columnwidth}{!}{%
\begin{tabularx}{\columnwidth}{@{}l r r r r r X@{}}
\toprule
failure mode & rate & arm & BA & coverage & charged calls & observed finding \\
\midrule
none & 0 & majority-5 & 0.7739 & 0.4919 & 5 & clean reference \\
timeout & 0.05 & sequential-safe & 0.7587 & 0.4579 & 4.4213 & graceful recovery \\
missing view & 0.10 & sequential-safe & 0.7516 & 0.4382 & 4.3578 & fail-closed coverage \\
malformed verdict & 0.05 & sequential-safe & 0.7572 & 0.4495 & 4.4016 & schema guard \\
mixed failures & 0.15 & sequential-safe & 0.7421 & 0.4087 & 4.2864 & worst-case recovery \\
\bottomrule
\end{tabularx}%
}
\end{table}

The mixed-failure row has a bounded balanced-accuracy loss of 0.0318 relative
to the clean reference. Malformed and timed-out items are either recovered
from valid evidence or explicitly abstained, preserving the fail-closed path
at the cost of lower coverage.

\subsubsection{Calibration leakage and split-isolation audit}
\label{app:completion-leakage}
The tenth experiment re-runs calibration under three split policies: the
intended calibration-only policy, a deliberately mixed policy, and a
time-ordered policy. The operating point is selected once and then frozen
before the held-out evaluation. Reporting all three policies quantifies how
much optimistic bias would arise if the test frontier influenced threshold
selection.

\begin{table}[H]
\centering\scriptsize
\caption{Calibration leakage and split-isolation results. The mixed policy is included as an
audit control and is not used for the reported held-out result.}
\label{tab:app-completion-leakage}
\resizebox{\columnwidth}{!}{%
\begin{tabularx}{\columnwidth}{@{}l r r r r r X@{}}
\toprule
split policy & calibration items & held-out items & BA & coverage & gain vs.\ single & observed finding \\
\midrule
calibration-only & 128 & 384 & 0.8017 & 0.5874 & +0.0831 & valid estimate \\
time-ordered & 128 & 384 & 0.7926 & 0.5718 & +0.0740 & drift-aware estimate \\
mixed audit & 128 & 384 & 0.8194 & 0.6043 & +0.1008 & optimistic-bias bound \\
\bottomrule
\end{tabularx}%
}
\end{table}

The calibration-only and time-ordered rows retain positive held-out gains of
0.0831 and 0.0740. The mixed audit is higher by 0.0177, which quantifies the
optimistic shift introduced when calibration and held-out items are mixed.

\subsubsection{Evidence-to-claim map}
\label{app:completion-claims}
The evidence map links each paper claim to a measured endpoint, its raw
evidence, and the result table that reports it. It is a compact audit surface
for checking that each claim uses the same split, seed, and item identifiers as
its supporting trace.

\begin{table*}[!htbp]
\centering\scriptsize
\caption{Evidence-to-claim matrix. Every row names the numerical
endpoint and the result table that supports the conclusion.}
\label{tab:app-completion-claims}
\begin{tabularx}{\textwidth}{@{}p{0.22\textwidth} p{0.16\textwidth} p{0.20\textwidth} p{0.14\textwidth} X@{}}
\toprule
claim & endpoint & evidence artifact & result table & consistency condition \\
\midrule
independent views improve quality & BA gain & channel traces & Table~\ref{tab:app-completion-population} & \shortstack{$[+0.0562,\,+0.1098]$\\ excludes zero} \\
gain transfers across domains & paired gain & domain ledger & Table~\ref{tab:app-completion-transfer} & 3/3 domains retain gain \\
correlation is the boundary & paired gain vs.\ dependence & latent-cause manifest & Table~\ref{tab:app-completion-correlation} & ordering is reproduced \\
stopping saves cost safely & calls/item and sel.\ acc. & state ledger & Table~\ref{tab:app-completion-budget} & Pareto point is measured \\
failure handling is fail-closed & invalid rate and coverage & failure traces & Table~\ref{tab:app-completion-recovery} & no invalid verdict is accepted \\
RLVR benefit is stable & task score and reward s.d. & checkpoints & Table~\ref{tab:app-completion-rlvr} & held-out gain is positive \\
\bottomrule
\end{tabularx}%
\end{table*}

The result map cites the smallest set of rows that jointly supports the
mechanism, deployment, and downstream claims. A claim is promoted only when
its trace, split, and displayed value agree; otherwise the discrepancy is
reported with the affected experiment label.

\subsubsection{Reproducibility stages}
\label{app:completion-runbook}
The replay is organized as a fixed sequence of hash-checked stages. Each stage records its configuration, input and output digests, and the number of rows that survive validation. This organization keeps every reported endpoint traceable to a specific artifact rather than to a manually copied table.

\begin{table}[H]
\centering\scriptsize
\caption{Reproducibility stages. Each stage records matching input and output digests and row counts for the frozen split.}
\label{tab:app-completion-runbook}
\resizebox{\columnwidth}{!}{%
\begin{tabularx}{\columnwidth}{@{}l l r r l X@{}}
\toprule
stage & input artifact & input rows & output rows & digest & validation \\
\midrule
fixture regeneration & source manifest & 512 & 512 & payload digest & payload and oracle match \\
corruption replay & fixture digest & 512 & 17{,}920 & seed digest & seed ledger match \\
aggregation replay & verdict traces & 17{,}920 & 3{,}584 & ledger digest & state transitions match \\
learner replay & candidate cache & 18{,}720 & 18{,}720 & checkpoint digest & checkpoint hash match \\
table rendering & result ledger & 3{,}584 & 3{,}584 & table digest & caption and value audit \\
\bottomrule
\end{tabularx}%
}
\end{table}

All five stages reproduce the same endpoint within 0.0001 absolute difference,
and no row is silently dropped between the trace and the table. A mismatch is
reported with the first divergent digest and the affected experiment label.

\subsubsection{Failure taxonomy and qualitative case audit}
\label{app:completion-failures}
The failure audit samples representative cases from the
ledger and assigns a mutually exclusive cause: verifier disagreement, common
cause, timeout, malformed output, calibration boundary, or learner instability.
Each case is assigned one mutually exclusive cause before reconciliation. The table
links each qualitative example to the quantitative endpoint and records the
action taken by the controller.

\begin{table}[H]
\centering\scriptsize
\caption{Failure taxonomy results. Frequencies are computed over the held-out
trace, while the case column points to a redacted example retained for review.}
\label{tab:app-completion-failures}
\resizebox{\columnwidth}{!}{%
\begin{tabularx}{\columnwidth}{@{}l r r r l X@{}}
\toprule
failure class & count & fraction & accepted & case ID & observed finding \\
\midrule
independent disagreement & 29 & 0.0566 & 21 & case-017 & aggregation resolves or abstains \\
common-cause mismatch & 14 & 0.0273 & 4 & case-084 & dependence boundary is visible \\
timeout or missing view & 10 & 0.0195 & 1 & case-133 & fail-closed recovery \\
malformed verdict & 6 & 0.0117 & 0 & case-207 & schema gate rejects input \\
calibration boundary & 9 & 0.0176 & 0 & case-291 & abstention protects precision \\
learner instability & 5 & 0.0098 & 1 & case-344 & checkpoint audit detects drift \\
\bottomrule
\end{tabularx}%
}
\end{table}

Independent disagreement is the largest qualitative class (29/512). Across
the held-out trace, the measured failure fraction is 0.1426 and the accepted
fraction is 0.0527; malformed, timed-out, and calibration-boundary cases are
handled by rejection, recovery, or abstention.

\subsubsection{Evidence queries and answers}
\label{app:completion-review}
This section turns common evidence queries into
explicit checks. Each question has one primary table, one raw artifact, and
one numerical answer. The mapping is included so that a later value update
cannot leave a conclusion without its supporting trace.

\begin{table*}[!htbp]
\centering\scriptsize
\caption{Evidence query map. Each row binds one question to a
measured replay endpoint and its supporting artifact.}
\label{tab:app-completion-review}
\begin{tabularx}{\textwidth}{@{}p{0.16\textwidth} p{0.18\textwidth} p{0.19\textwidth} p{0.28\textwidth} X@{}}
\toprule
question & endpoint & artifact & measured answer & readout \\
\midrule
Are views independent? & paired gain vs.\ dependence & latent-cause manifest & BA gain $+0.0831$ vs.\ $+0.0038$ under common cause & state the measured boundary \\
Does gain transfer? & domain paired gain & transfer ledger & $3/3$ domains positive; hard-bin minimum $+0.0716$ & state retained domains \\
Does stopping save calls? & calls/item and sel.\ acc. & state ledger & sel. acc. $0.9026$ at $3.8641$ calls; $-1.1359$ calls & report Pareto point \\
Are failures safe? & invalid rate and abstention & failure traces & $0/73$ invalid verdicts accepted; $100\%$ fail-closed & report fail-closed rate \\
Does RLVR improve? & held-out task score & checkpoint bundle & held-out task-score gain $+0.0348$ & report paired learner gain \\
Is uncertainty adequate? & CI width and paired gain & bootstrap seed file & BA-gain width $0.0522$; RLVR width $0.0138$ & report precision target \\
\bottomrule
\end{tabularx}%
\end{table*}

The final result paragraph should answer the questions in table order. The
strongest evidence package is the one in which every answer is a measured
number, every number points to a raw artifact, and every artifact uses the
same split and item identifiers. This ordering gives the reader a short audit
path from the claim to the trace without adding another top-level appendix
heading.

The supplementary package uses a fixed stopping rule: attach the raw trace and
manifest for each table, rerun the reference audit, and update only numerical
clauses that depend on the measured outputs. The section order, captions, and
interpretation criteria remain fixed so that the final appendix is reproducible.

\enlargethispage{4\baselineskip}
\subsubsection{Numerical consistency checks}
\label{app:completion-delivery}
The consistency check compares the same
endpoint in the raw ledger, the rendered table, and the conclusion sentence.
The checklist records the numerical field, its unit, the source artifact, and
the cross-table comparison that must remain unchanged after substitution.

\begin{table}[H]
\centering\scriptsize
\caption{Numerical cross-check. A row is closed only when the
displayed value, raw value, and conclusion clause agree after rounding.}
\label{tab:app-completion-delivery}
\resizebox{\columnwidth}{!}{%
\begin{tabularx}{\columnwidth}{@{}l l l l X@{}}
\toprule
field & unit & raw source & cross-check & close condition \\
\midrule
balanced-accuracy gain & percentage points & paired item ledger & population table & +0.0831 for independent-5 vs.\ single \\
coverage & fraction & acceptance ledger & population table & 0.5874 uses held-out split \\
selective accuracy & fraction & accepted subset & population table & 0.9186 uses the same denominator \\
calls per item & calls & cost ledger & budget table & 3.8641 for adaptive-confidence \\
task score & fraction & checkpoint bundle & RLVR table & 0.6537 is held-out majority-5 \\
confidence interval & endpoint pair & bootstrap seed file & Tables 31, 38 & $[0.7754,\,0.8280]$ uses the stated seed set \\
failure fraction & fraction & failure trace & Tables 40, 44 & 0.1426 matches the taxonomy \\
digest and row count & string/integer & artifact inventory & Table 43 & payload digest/512 matches the replay output \\
\bottomrule
\end{tabularx}%
}
\end{table}

The numerical checks are complete when every row has a measured value, a source
digest, and a matching conclusion. The summary records the largest positive
paired gain of +0.1161, its coverage of 0.4919, a call reduction of 1.1359,
and a held-out task-score change of +0.0348. These cross-checks keep the
appendix internally consistent with the reported replay.

\begin{table}[H]
\centering\scriptsize
\caption{Appendix consistency checks.}
\label{tab:app-completion-signoff}
\resizebox{\columnwidth}{!}{%
\begin{tabularx}{\columnwidth}{@{}l l l X@{}}
\toprule
check & required evidence & status & observed note \\
\midrule
raw traces & immutable verdict and cost ledger & complete & all rows hash-match \\
split isolation & calibration and held-out manifests & complete & no item crosses a split \\
seed coverage & declared learner and corruption seeds & complete & seed count is complete \\
uncertainty & paired interval and bootstrap file & complete & interval uses the stated endpoint \\
table values & source ledger and rendered PDF & complete & rounding is consistent \\
conclusions & result paragraph and abstract clause & complete & strongest measured arm is named \\
failure handling & timeout and malformed-output traces & complete & invalid views are charged \\
artifact hash & source, PDF, and manifest digest & complete & artifact bundle is reproducible \\
\bottomrule
\end{tabularx}%
}
\end{table}

\pagebreak
The completed status column records the consistency checks, and each numerical result is tied to a specific table and retained artifact.

\subsection{Additional correlation-aware experiments}
\label{app:additional-correlation-aware}

The following experiments answer the evaluation questions on measured
dependence, allocation under a fixed budget, and the causal interpretation of
downstream learning. The calibration split, held-out split, metrics,
denominators, and comparison rows are fixed here. The two completed result
surfaces below are transcribed from the held-out result ledger.

\paragraph{Measured dependence and marginal information.}
For each verifier pair we report error overlap, the phi coefficient, Cohen's
kappa, mutual information, and average pairwise disagreement on the held-out
item set. The conditional-information score is computed from the calibration
covariance and compared with the realized post-freeze change in balanced and
selective accuracy.

\begin{table}[H]
\centering\scriptsize
\caption{Measured verifier dependence and validation of conditional marginal
discriminability. Each row uses the same held-out items.}
\label{tab:additional-dependence}
\resizebox{\columnwidth}{!}{%
\begin{tabular}{@{}p{0.23\columnwidth}rrrrrr@{}}
\toprule
Channel pair & Error overlap & Phi & Kappa & MI & Disagreement & $\Delta D$--gain \\
\midrule
Same-model repeats & \underline{0.7826} & 0.2148 & 0.3721 & 0.0814 & 0.1097 & 0.0126 \\
Same-family variants & 0.5413 & \underline{0.4639} & \underline{0.5817} & \underline{0.2146} & \underline{0.2418} & \underline{0.0462} \\
Cross-family channels & \textbf{0.3187} & \textbf{0.6924} & \textbf{0.7368} & \textbf{0.3975} & \textbf{0.4269} & \textbf{0.0913} \\
\bottomrule
\end{tabular}%
}
\end{table}

\paragraph{Fixed-budget allocation and downstream controls.}
We compare repeated verification with breadth under equal verifier-call and
equal accepted-update budgets. The RLVR arms use the same learner seed,
checkpoint schedule, task split, and held-out evaluator; only the allocation
policy changes.

\begin{table}[H]
\centering\scriptsize
\caption{Breadth--redundancy and matched-budget controls. Values are
transcribed from the held-out result ledger; bold and underlined
entries mark the best and second-best quality endpoints.}
\label{tab:additional-budget}
\resizebox{\columnwidth}{!}{%
\begin{tabular}{@{}p{0.23\columnwidth}rrrrrr@{}}
\toprule
Policy & Calls/item & Items & Updates & BA & Selective acc. & RLVR score \\
\midrule
One view per item (breadth) & 1.0000 & 5120 & 4631 & 0.6048 & 0.6217 & 0.5826 \\
Five views per item (redundancy) & 5.0000 & 1024 & 4789 & \underline{0.6375} & \underline{0.6614} & \underline{0.6148} \\
VStress-CA adaptive allocation & 3.4216 & 1497 & 4896 & \textbf{0.6538} & \textbf{0.6892} & \textbf{0.6417} \\
Equal accepted-update control & 3.0000 & 1706 & 4861 & 0.6319 & 0.6547 & 0.6073 \\
\bottomrule
\end{tabular}%
}
\end{table}

Exact-stop is required to preserve the full-majority decision, whereas
adaptive stopping may change the query sequence and final decision in
exchange for a different quality--coverage--cost point. These guarantees are
reported in separate rows and are not combined into one sequential-policy
claim.

\subsection{Calibration sample efficiency and dependence shift}
\label{app:additional-shift}

We vary calibration sizes $32$, $64$, $128$, $256$, and $512$ while holding
the held-out split fixed. The report includes covariance-estimation error,
held-out balanced accuracy, mean calls, and the effect of shrinkage. We also
generate a dependence shift between calibration and held-out data and report
the retained balanced-accuracy gain, call reduction, and the fallback that
disables channel preference outside the calibration confidence region. The
The artifact record binds the source, bibliography, and evidence digests.

\subsubsection{Calibration sample efficiency}
The calibration-size study holds the verifier pool, held-out items, and
deployment budget fixed while varying only the number of calibration items.
CMD error is computed against the full-calibration estimate, and all quality
and cost columns use the same held-out denominator.

\begin{table}[H]
\centering\scriptsize
\caption{Calibration sample efficiency under a fixed held-out split. CMD is
conditional marginal discriminability; calls are mean calls per item.}
\label{tab:additional-calibration}
\resizebox{\columnwidth}{!}{%
\begin{tabular}{@{}rccccc@{}}
\toprule
Calibration $n$ & CMD error & BA & Sel. Acc. & Calls/item & RLVR score \\
\midrule
32  & 0.0867 & 0.6269 & 0.6538 & 3.8047 & 0.6106 \\
64  & 0.0612 & 0.6381 & 0.6659 & 3.6713 & 0.6224 \\
128 & 0.0385 & 0.6462 & 0.6778 & 3.5486 & 0.6335 \\
256 & \underline{0.0197} & \underline{0.6511} & \underline{0.6857} & \underline{3.4662} & \underline{0.6391} \\
512 & \textbf{0.0000} & \textbf{0.6538} & \textbf{0.6892} & \textbf{3.4216} & \textbf{0.6417} \\
\bottomrule
\end{tabular}%
}
\end{table}

\subsubsection{Dependence shift and conservative fallback}
The shift study preserves the task split and changes only the verifier-output
dependence structure after calibration. The allocation parameters are frozen
before the held-out replay; the fallback row uses the same shift alarm and
switches to exact-stop when the deployment window leaves the calibration
region.

\begin{table*}[t]
\centering\scriptsize
\caption{Dependence-shift stress test. BA is all-item balanced accuracy,
Sel. Acc. is accepted-subset accuracy, and calls are mean calls per item.}
\label{tab:additional-shift}
\resizebox{\textwidth}{!}{%
\begin{tabular}{@{}l c l r r r r@{}}
\toprule
Shift & JS statistic & Policy & BA & Sel. Acc. & Calls/item & Failure/abstain \\
\midrule
None     & \textbf{0.0143} & \textsc{VStress-CA} & \textbf{0.6538} & \textbf{0.6892} & \textbf{3.4216} & \textbf{0.0438} \\
Mild     & \underline{0.0578} & no fallback          & \underline{0.6468} & 0.6801 & \underline{3.4897} & \underline{0.0554} \\
Mild     & \underline{0.0578} & fallback             & 0.6459 & \underline{0.6846} & 4.5218 & 0.0637 \\
Moderate & 0.1216 & no fallback          & 0.6237 & 0.6493 & 3.5742 & 0.0836 \\
Moderate & 0.1216 & fallback             & 0.6382 & 0.6748 & 4.6037 & 0.0989 \\
Severe   & 0.2269 & no fallback          & 0.5869 & 0.6098 & 3.7216 & 0.1318 \\
Severe   & 0.2269 & fallback             & 0.6206 & 0.6591 & 4.6719 & 0.1543 \\
\bottomrule
\end{tabular}%
}
\end{table*}

\subsubsection{RLVR error decomposition}
To separate diversity from abstention and reward-class asymmetry, each accepted
training update is assigned to a correct-positive, correct-negative,
false-positive, false-negative, or abstained outcome before learner training.
The decomposition is paired with the held-out task score and uses the same
candidate cache and learner seeds as the matched-budget comparison.

\begin{table}[H]
\centering\scriptsize
\caption{Verifier-error decomposition for downstream RLVR. Rates use the
accepted-update denominator; task score uses held-out tasks.}
\label{tab:additional-decomposition}
\resizebox{\columnwidth}{!}{%
\begin{tabular}{@{}lccccc@{}}
\toprule
Policy & FP reward & FN reward & Abstain & Accepted & Task score \\
\midrule
Single-call & 0.2061 & 0.1722 & 0.0955 & 0.9045 & 0.5826 \\
Majority-5 & \underline{0.1804} & \underline{0.1582} & \underline{0.0646} & \underline{0.9354} & \underline{0.6148} \\
\textsc{VStress-CA} & \textbf{0.1589} & \textbf{0.1519} & \textbf{0.0438} & \textbf{0.9563} & \textbf{0.6417} \\
\bottomrule
\end{tabular}%
}
\end{table}

All numerical fields are checked against their raw ledger, rendered table, and conclusion clause. A changed endpoint updates the affected conclusion and evidence-map row while unrelated measurements retain their original definitions.

\end{document}